%% file: PaperForCamera.tex
\documentclass[10pt,twocolumn,letterpaper]{article}

\usepackage[pagenumbers]{wacv} 

\usepackage{graphicx}
\usepackage{amsmath}
\usepackage{amssymb}
\usepackage{booktabs}
\usepackage{colortbl}
\usepackage{xcolor}
\usepackage{color}
\usepackage{pifont}
\usepackage{colortbl}
\usepackage{booktabs}
\usepackage{multirow}
\definecolor{OliveGreen}{rgb}{0,0.6,0}
\definecolor{BrickRed}{rgb}{0.8,0.25,0.33}
\definecolor{baselinecolor}{gray}{.9}
\definecolor{backcolor}{RGB}{232, 242, 255}
\definecolor{improvecolor}{RGB}{112, 173, 71}
\newcommand{\gou}{{\color{OliveGreen}\ding{51}}}
\newcommand{\cha}{{\color{BrickRed}\ding{55}}}

\usepackage{pgffor}
\newcommand{\Repeat}[2]{
    \foreach \n in {1,...,#1}{#2}
}

\usepackage[accsupp]{axessibility} 

\usepackage[pagebackref,breaklinks,colorlinks]{hyperref}

\usepackage[capitalize]{cleveref}
\crefname{section}{Sec.}{Secs.}
\Crefname{section}{Section}{Sections}
\Crefname{table}{Table}{Tables}
\crefname{table}{Tab.}{Tabs.}

\def\wacvPaperID{***} 
\def\confName{WACV}
\def\confYear{2025}

\begin{document}

\title{Automated Evaluation of Large Vision-Language Models on Self-driving \\ Corner Cases}

\author{
Kai Chen$^{1}$\thanks{\ Equal contribution. $^{\dagger}$ Corresponding authors. \\ Contact: \texttt{jiayushenyang@gmail.com}}\quad
Yanze Li$^{2*}$\quad
Wenhua Zhang$^{2*}$\quad 
Yanxin Liu$^{2}$\quad
Pengxiang Li$^{2}$\quad 
Ruiyuan Gao$^{3}$ \\
Lanqing Hong$^{4\dagger}$\quad 
Meng Tian$^{4}$\quad 
Xinhai Zhao$^{4}$\quad  
Zhenguo Li$^{4}$  \\
Dit-Yan Yeung$^{1}$\quad 
Huchuan Lu$^{2}$\quad 
Xu Jia$^{2\dagger}$
\\
$^{1}$Hong Kong University of Science and Technology\quad
$^{2}$Dalian University of Technology
\\
$^{3}$The Chinese University of Hong Kong\quad
$^{4}$Huawei Noah's Ark Lab
\\
\small\url{https://coda-dataset.github.io/coda-lm/}
}

\maketitle


\input{sec/0_abstract}
\input{sec/1_introduction}
\input{sec/2_related_works}
\input{sec/3_CODALLM_Dataset}
\input{sec/4_evaluation}
\input{sec/5_Evaluation_Results}
\input{sec/6_conclusion}

\paragraph{Acknowledgments.}
We gratefully acknowledge supports of MindSpore, CANN (Compute Architecture for Neural Networks) and Ascend AI Processor used for this research.
The research was partially supported by the National Natural Science Foundation of China (grants Nos. 62472065, U23B2010, and 62106036).
This research has been made possible by funding support from the Research Grants Council of Hong Kong through the Research Impact Fund project R6003-21.

{\small
\bibliographystyle{ieee_fullname}
\bibliography{egbib}
}

\clearpage
\input{sec/7_appendix}

\end{document}

%% file: sec/0_abstract.tex
\begin{abstract}
Large Vision-Language Models (LVLMs) have received widespread attentions for advancing the interpretable self-driving.
Existing evaluations of LVLMs primarily focus on multi-faceted capabilities in natural circumstances, lacking automated and quantifiable assessment for self-driving, let alone the severe road corner cases.
In this work, we propose \textbf{CODA-LM}, the very first benchmark for the automatic evaluation of LVLMs for self-driving corner cases. 
We adopt a hierarchical data structure and prompt powerful LVLMs to analyze complex driving scenes and generate high-quality pre-annotations for the human annotators, while for LVLM evaluation, we show that using the text-only large language models (LLMs) as judges reveals even better alignment with human preferences than the LVLM judges.
Moreover, with our CODA-LM, we build \textbf{CODA-VLM}, a new driving LVLM surpassing all open-sourced counterparts on CODA-LM.
Our CODA-VLM performs comparably with GPT-4V, even surpassing GPT-4V by \textbf{+21.42\%} on the regional perception task.
We hope CODA-LM can become the catalyst to promote interpretable self-driving empowered by LVLMs.

\end{abstract}

%% file: sec/1_introduction.tex
\section{Introduction}
\label{sec:intro}
Large Vision-Language Models (LVLMs)~\cite{Liu2023VisualIT,gpt4v,gou2023mixture,chen2024emova} have attracted increasing attention, primarily due to their remarkable visual reasoning abilities, which are of paramount importance~\cite{hu2023planning,sima2023drivelm} for the autonomous driving.
Traditional self-driving systems use a modular design, integrating various modules including perception, prediction, and planning to handle complicated road scenarios, 
which, however, are still inadequate to generalize in the open domain, especially for the severe real-world \textit{corner cases}~\cite{li2022coda}. 
In this paper, we primarily consider \textit{object-level corner cases}\footnote{We adopt the definition of object-level corner case in~\cite{breitenstein2021corner}.}, including both the \textit{instances of novel categories} and \textit{novel instances of common categories}~\cite{li2022coda}.


\begin{figure}[t]
\centering
\includegraphics[width=\linewidth]{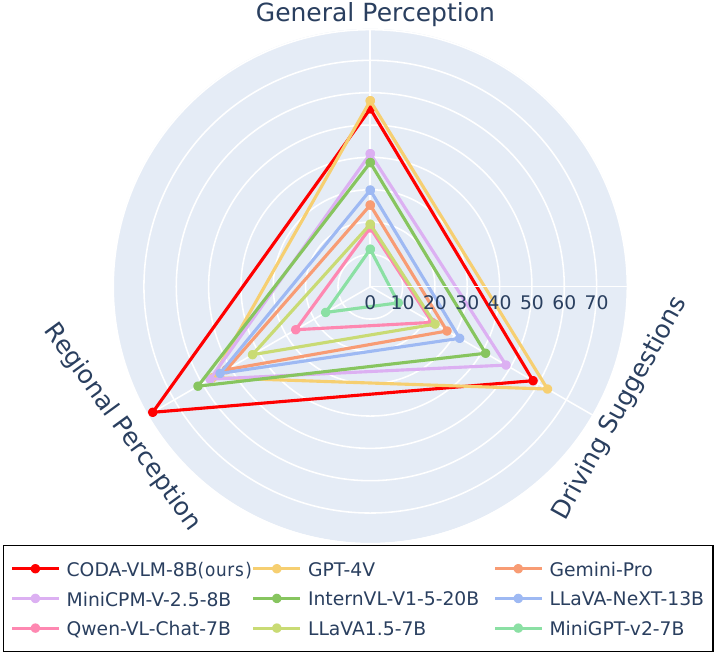}
\caption{
\textbf{Comparison among open-sourced and commercial LVLMs on CODA-LM}.
CODA-LM provides the very first automated and quantifiable evaluation of LVLMs on road corner cases.
}
\label{fig:model radar}
\end{figure}


LVLMs, on the other hand, with their extensive world knowledge and reasoning capability, have the potential to overcome these severe challenges. 
A preliminary study~\cite{wen2023road} has revealed the capability of powerful LVLMs~\cite{gpt4v} in handling the road corner cases, 
where the samples are selected from CODA~\cite{li2022coda}, the largest real-world corner case dataset, to prompt GPT-4V.
Although effective, their evaluation relies on redundant manual inspections, hindering the scalability of larger-scale LVLM evaluation for self-driving.


\begin{table*}[t]
\centering
\scalebox{1.}{
\setlength{\tabcolsep}{2mm}{
\begin{tabular}{l|c c | c c c@{}}
\toprule
Dataset & Multimodal  & Corner & General Perception & Regional Perception &  Driving Suggestion  \\
\midrule
CODA~\cite{li2022coda} & \cha & \gou & \gou & \gou & \cha \\
StreetHazards~\cite{hendrycks2019benchmark} & \cha & \gou & \gou & \gou & \cha \\
\midrule
nuScenes-QA~\cite{qian2023nuscenes} & \gou & \cha & \gou & \cha & \cha \\
BDD-X~\cite{kim2018textual} &\gou & \cha & \gou & \cha & \cha \\
DRAMA~\cite{malla2023drama}  & \gou & \cha & \gou & \gou & \gou \\
DriveLM~\cite{sima2023drivelm}  &\gou & \cha & \gou & \gou & \gou \\
\midrule
\rowcolor{backcolor}
\textbf{CODA-LM (ours)}  & \gou & \gou & \gou & \gou & \gou \\
\bottomrule
\end{tabular}
}}
\vspace{-2mm}
\caption{\textbf{Comparison between CODA-LM and existing datasets}.
CODA-LM is the first large-scale multimodal road corner case dataset for interpretable autonomous driving with an automatic and hierarchical evaluation framework.
}
\vspace{-3mm}
\label{tab:dataset_comparison}
\end{table*}


In this paper, we propose the \textbf{CODA-LM}, the very first benchmark for the automated and systematic evaluation of LVLMs on the self-driving corner cases. 
Following Wen~\etal~\cite{wen2023road}, we utilize the corner cases from CODA and collect question-answering annotations of three distinct tasks including \textit{general perception}, \textit{regional perception}, and \textit{driving suggestions}.
To obtain high-quality pre-annotation, we design a hierarchy data structure to help GPT-4V better analyze complex road scenes and capture all necessary obstacles.
The structured responses are then converted to coherent texts, which are then verified by human annotators.
Different from the existing LVLM benchmarks~\cite{li2024llavanext-strong}, we show the necessity of using the text-only LLMs~\cite{chatgpt} as ``judges'' for automated evaluation of LVLMs on CODA-LM, which reveals
a stronger consistency with humans than LVLM judges~\cite{gpt4v}.
Moreover, we propose \textbf{CODA-VLM}, a novel driving LVLM achieving the state-of-the-art among all open-sourced LVLMs on CODA-LM, even surpassing GPT-4V on the regional perception task by \textbf{+21.42\%}.

The main contributions of this work contain three parts:
\begin{enumerate}
  \item We propose \textbf{CODA-LM}, the \textbf{very first} LVLM benchmark for the automatic and systematic evaluation of LVLMs on road corner cases.

  \item We demonstrate that text-only LLMs can serve as powerful judges to evaluate LVLMs, revealing a stronger consistency with the human judgments even compared with LVLM judges.

  \item We comprehensively assess the performance of existing LVLMs on self-driving corner cases, and construct \textbf{CODA-VLM}, a new driving LVLM comparable with GPT-4V on CODA-LM, surpassing all open-sourced counterparts on both perception and suggestions.
\end{enumerate}

%% file: sec/2_related_works.tex
\section{Related Work}
\paragraph{LVLM evaluation} primarily focuses on natural image spaces.
MME~\cite{fu2023mme} introduces manually designed question-answering pairs to measure both perception and cognition capabilities on a total of 14 sub-tasks.
MMBench~\cite{liu2023mmbench} employs GPT-4 to transform free-form predictions into pre-defined multiple-choice questions and introduces the CircularEval strategy for a more robust evaluation. 
The SEED-Bench-2~\cite{li2023seed} adopts a similar format with MMBench but extends over 27 dimensions, evaluating LVLMs' abilities in image and text comprehension and interleaved image-text understanding and generation tasks.
Auto-Bench~\cite{ji2023large} generates question-answer-reasoning triplets using LLMs \cite{touvron2023llama,chen2023gaining,liu2024mixture,gou2024eyes} as evaluation data. 
Tri-HE~\cite{wu2024unified}, instead, focuses on LVLM hallucination with a unified evaluation framework.
All the evaluation benchmarks above rely on the rigid, manually curated datasets of natural images, and thus, difficult to apply for complicated driving scenarios.

\paragraph{Autonomous driving datasets.}
The NuScenes-QA~\cite{qian2023nuscenes} manually constructs 460K question-answer pairs based on the object attributes and relationships among objects in scene graphs. 
BDD-X~\cite{kim2018textual} focuses on the behavior of the ego car and provides corresponding reasons. 
While both datasets concentrate on general perception, 
DRAMA~\cite{malla2023drama} and DriveLM~\cite{sima2023drivelm} further consider regional perception and driving suggestions.
DRAMA identifies the most critical targets and offers the corresponding advice, while DriveLM promotes end-to-end autonomous driving understanding through the usage of graph-structured question-answer pairs. 
Self-driving systems often fail in corner cases, leading to severe accidents. StreetHazards~\cite{hendrycks2019benchmark} is a synthesized dataset where corner cases are simulated via graphics. 
CODA~\cite{li2022coda} is a real-world road corner case dataset with 10K driving scenes, spanning more than 40 classes. 
As in Tab.~\ref{tab:dataset_comparison}, the existing corner case datasets lack language modality, while vision-language datasets don't cover road corner cases. 
Thus, we propose CODA-LM, the first large-scale multimodal road corner case dataset for self-driving with a hierarchical automatic evaluation framework.

%% file: sec/3_CODALLM_Dataset.tex

\begin{figure}[t]
  \centering
  \includegraphics[width=\linewidth]{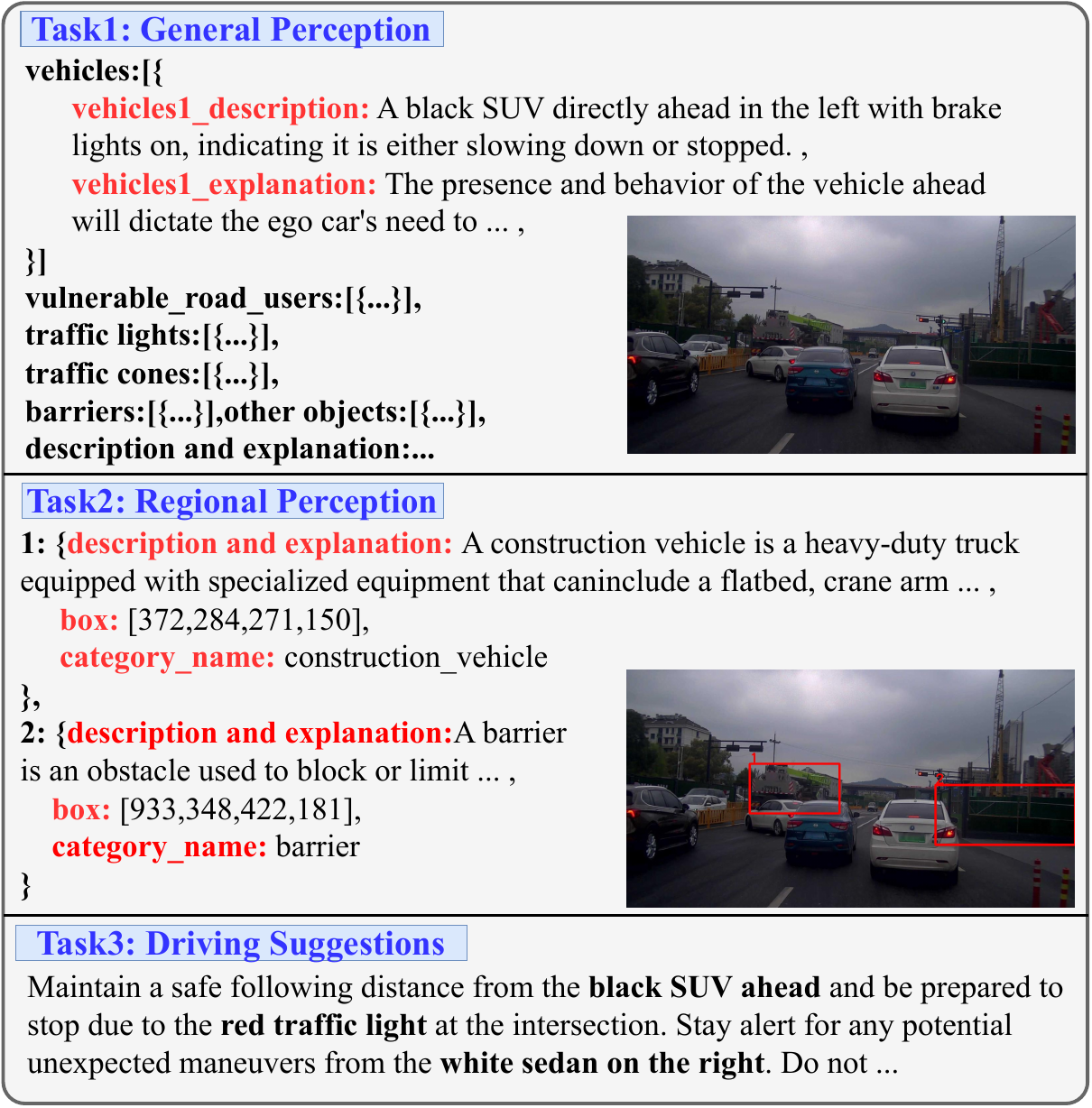}
  \caption{
  \textbf{Task hierarchy of our CODA-LM}, including \textit{general perception} (up), \textit{regional perception} (middle), and \textit{driving suggestions} (bottom), respectively.
  }
  \label{fig:dataset example}
\end{figure}


\section{CODA-LM Dataset}\label{sec:blind}
Based on the road corner cases from CODA~\cite{li2022coda}, our CODA-LM comprises 9,768 real-world driving scenarios with 41,722 textual annotations for critical road entities and 21,537 annotations for road corner cases. 
Critical road entities affecting self-driving decision-making are categorized into seven distinct groups, including \textit{vehicles}, \textit{vulnerable road users (VRUs)}, \textit{traffic signs}, \textit{traffic lights}, \textit{traffic cones}, \textit{barriers}, and \textit{other objects} (\eg, animals and traffic islands).
As illustrated in Fig.~\ref{fig:dataset example}, our CODA-LM involves a task hierarchy with three principal tasks, including the \textit{general perception}, \textit{regional perception}, and \textit{driving suggestion}, as detailed in Sec.~\ref{sec:general_perception}-\ref{sec:suggestion} separately.
Such a systematic task hierarchy requires LVLMs to understand complex driving situations, providing a comprehensive assessment of \textbf{interpretable} self-driving agents empowered by LVLMs.


\subsection{General Perception}\label{sec:general_perception}
The foundational aspect of the general perception task lies in a comprehensive understanding of critical road key entities in driving scenarios, including their appearance, location, and reasons why they influence the driving behaviors of our ego car. 
This task is pivotal in evaluating LVLMs' proficiency in interpreting complex interactive scenes, mirroring the perception process in self-driving. 
Moreover, to comprehensively evaluate LVLMs' performance in different environments, we classify the images based on the time and weather conditions, including \textit{night} and \textit{daytime} scenes for time conditions, as well as \textit{clear}, \textit{cloudy}, and \textit{rainy} circumstances for the weather conditions.


\subsection{Regional Perception}\label{sec:regional_perception}
The regional perception task measures LVLMs' capabilities to understand corner case objects when provided with specific bounding boxes, which involves describing objects within the given bounding boxes and explaining why they would influence self-driving behavior.
The establishment of regional perception is based on a core realization \cite{han2021soda10m} that the ability to accurately localize corner cases is crucial for enhancing the overall system's robustness in the practical application of autonomous driving. 
These scenarios often contain complicated or unusual elements that traditional models might overlook or struggle to interpret correctly, such as \textit{unique traffic signs}, \textit{pedestrians with abnormal behavior}, and \textit{atypical road conditions}. By specifically focusing on these cases, we can gain a comprehensive understanding of LVLMs' ability to comprehend corner cases.


\subsection{Driving Suggestions}\label{sec:suggestion}
The driving suggestions task aims to evaluate the capability of LVLMs in formulating driving advice, a critical component for interpretable self-driving.
This task is closely related to the planning process of autonomous driving, requiring the model to provide the optimal driving suggestions for the ego car after correctly perceiving the general and regional aspects of the current driving environment. 
Via the construction of the driving suggestions task, we can deeply evaluate the performance of LVLMs in formulating effective driving strategies.

%% file: sec/4_evaluation.tex

\begin{figure*}[t]   
  \centering
  \includegraphics[width=\linewidth]{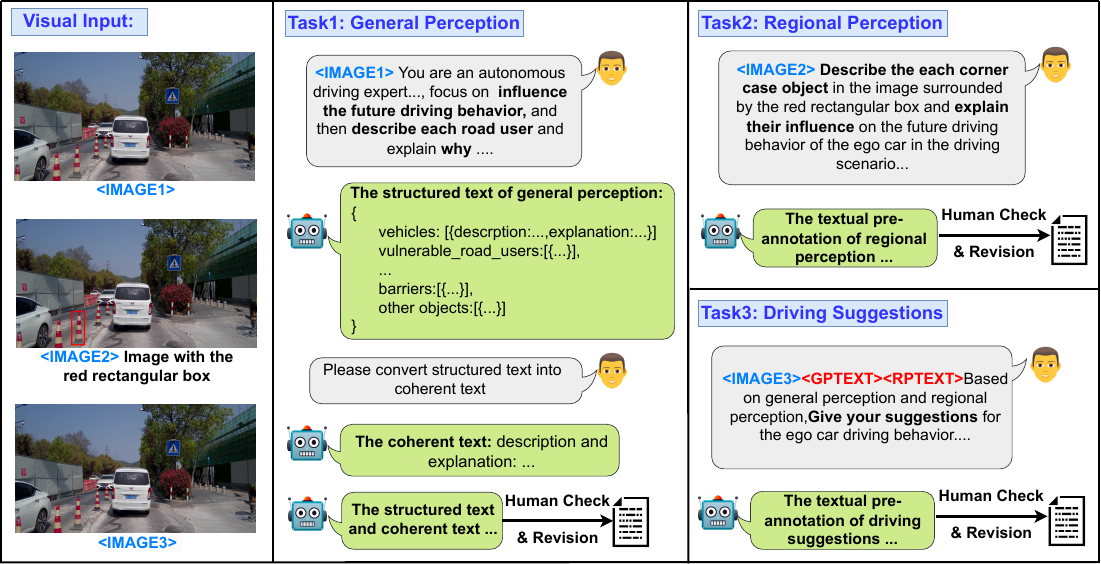}
  \caption{
  \textbf{Overview of CODA-LM construction.}
  We design a hierarchical data structure in the JSON format to guide GPT-4V to better understand complicated driving scenes and generate high-quality pre-annotations for human annotators to conduct further verification and revision.
  \textcolor{red}{\texttt{<GPTEXT>}} and \textcolor{red}{\texttt{<RPTEXT>}} refer to the revised answer from the general perception and the regional perception, respectively.
  }
  \vspace{-3mm}
  \label{fig:data collection}
\end{figure*}


\section{CODA-LM Construction}\label{sec:construction}
\subsection{Data Collection}\label{sec:collection}
\paragraph{Overview.}
For each task introduced in Sec.~\ref{sec:blind}, we meticulously design prompts to guide GPT-4V
\footnote{https://chatgpt.ust.hk} 
to generate high-quality textual pre-annotations based on visual information, as provided in Figs.~\ref{fig:gen_pr} and~\ref{fig:reg_pr}.
We start by constructing a hierarchical data structure in the JSON format (detailed in the following) to guide GPT-4V for better scene understanding of complex road scenes, categorizing the critical road entities into seven classes.
Each entity is detailedly described, explaining how they affect the driving behavior of the ego car.
After obtaining the GPT-4V responses for both the general and regional perceptions, we combine these with the corresponding road image to form a composite context for the GPT-4V to generate the driving suggestions.
Finally, we ask human annotators to verify and revise the pre-annotations.
The construction pipeline is shown in Fig.~\ref{fig:data collection}.

\paragraph{Hierarchical text structure for general perception.}
To conduct precise perception and even driving suggestions, it is essential to recognize all road obstacles. 
However, if directly prompted with plain texts, we notice that GPT-4V suffers from 1) \textit{entity ignorance}: GPT-4V tends to focus on the salient objects while ignoring the insignificant obstacles.
2) \textit{element ignorance}: when prompted with plain texts, GPT-4V might describe road entities without explaining why it affects the ego car or vice versa.

Thus, as in Fig.~\ref{fig:data collection} (middle), we design a hierarchy data structure in the JSON format from \textit{categories} to \textit{objects} and ultimately \textit{data elements}.
GPT-4V is guided to first recognize objects of every single \textit{category} separately, and ``fill in'' \textit{description} and \textit{explanation} of each object.
We then prompt GPT-4V again to convert structured texts to coherent natural languages and serve as the final pre-annotations.
As in Tab.~\ref{tab:human consis reference}, the \textbf{``structure-coherence''} pipeline achieves significant consistency with humans.

\vspace{-1mm}
\paragraph{Visual prompts for regional perception.}
We consider two manners to convert bounding boxes as the inputs for LVLMs, 
1) \textit{visualization}: suggests marking the targets with red rectangle boxes on the original images, as in Fig.~\ref{fig:data collection} (left).
2) \textit{grounding}: uses normalized coordinates (top-left and bottom-right corners) in text prompts to locate the target, similarly with LLaVA~\cite{Liu2023VisualIT}.
As in Tab.~\ref{tab:ablation_region_data}, visualization with red rectangles reveals significantly better empirical results, which is considered as the default vision prompts.

\vspace{-1mm}
\paragraph{Human verification and revision} is ultimately adopted to guarantee the correctness of our CODA-LM annotations.
For convenience, we construct a labeling tool GUI based on Gradio~\cite{abid2019gradio}, as in Fig.~\ref{fig:label_tool}, followed by the ethics review.

\paragraph{Data split.}
We separate 4,884 scenes as the training set, with 4,384 data samples as the validation set and the remaining 500 samples as the test set to construct the CODA-LM benchmark as in Tab.~\ref{tab:comparison} for a comprehensive comparison among LVLMs.

\subsection{Evaluation Framework}\label{sec:evaluation}
\paragraph{Unsatisfactory LVLM judges.}
LMSYS~\cite{zheng2024judging} shows the feasibility of using GPT-4 as judges to evaluate the intelligent chat assistants by giving a 1-10 score, revealing high consistency with human assessment.
Inspired by that, we start with a preliminary attempt by using \textit{LVLM judges} (\eg, GPT-4V) to evaluate various LVLMs, which, however, merely obtains a human consistency of around 70\% for all three tasks, as shown in Tab.~\ref{tab:human consis}.

We assume that this is probably due to the unsatisfactory instruction-following ability of GPT-4V, which cannot always respond in the required format~\cite{bai2023touchstone}.
Meanwhile, GPT-4V still lacks the multimodal in-context learning ability, making few-shot evaluation indispensable in complex and varied autonomous driving scenarios.

\paragraph{Text-only LLM as LVLM judges.}
In this paper, we propose to adopt \textit{text-only LLMs} (\eg, GPT-4) 
as judges to evaluate LVLMs on driving scenarios.
Given the reference ground truths and few-shot ICL examples, GPT-4 is instructed to evaluate the correctness of model responses with a score ranging from 1 to 10.
The average score of the whole evaluation set serves as the final \texttt{Text-Score}.
We provide the evaluation prompts and ICL examples in Figs.~\ref{fig:eva_pr} to~\ref{fig:low_few_shot}.
As shown in Tab.~\ref{tab:human consis}, the text-only GPT-4 judge evaluates more consistently with human judgments than the GPT-4V judge.

\paragraph{Potential bias and hallucination}
To revise that, 
we ask the human annotators to verify and revise the evaluation results given by GPT-4 and finally report results in Tab.~\ref{tab:comparison}.

\paragraph{Evaluation criteria} of the general perception include \textit{accuracy}, \textit{hallucination penalty}, and \textit{consistency}.
Accuracy evaluates how well LVLMs match with reference ground truths, while the hallucination penalty suggests that LVLMs should not mention entities not collected in the reference, which, otherwise, should be penalized when computing scores.
Consistency focuses on the relationship between the object description and the explanation of why it affects the ego car.
For driving suggestions, the criteria focus on the \textit{rationality}, \textit{relevance}, and \textit{detail level} of driving suggestions generated by LVLMs. 
Especially for driving suggestions, we require responses to be specific and actionable, rather than vague or overly broad.
Prompts are listed in Fig~\ref{fig:eva_pr}.

\paragraph{Evaluation metrics.}
As previously introduced, we utilize the \texttt{Text-Score}~\cite{zheng2024judging} given by text-only GPT-4 judge as the primary evaluation metrics for all three tasks.
We further explore the usage of traditional text-generation evaluation metrics as in Tab.~\ref{tab:ablation_suggest}, which, however, cannot well differentiate the capabilities of various LVLMs under complicated self-driving scenarios.


\subsection{CODA-VLM}
In this section, we explore improving the performance of LVLM models on road corner cases from the perspectives of both the visual representation and knowledge transfer and construct our \textbf{CODA-VLM}, a novel driving LVLM achieving state-of-the-art recognition and planning performance on autonomous driving scenarios.

\paragraph{Knowledge transfer.}
To acquire more comprehensive pre-training knowledge, we use the LLaVA-Llama-3-8B-v1.1 developed by Xtuner\footnote{https://github.com/InternLM/xtuner} as our baseline, which follows the basic architecture of LLaVA1.5~\cite{liu2023improved}, while replacing the LLM with LLaMA3-8B\footnote{https://huggingface.co/meta-llama/Meta-Llama-3-8B}, and performing modality alignment and instruction fine-tuning on a larger dataset. 
Based on that, we inject knowledge specific to driving scenarios via instruction fine-tuning. 
Specifically, we organize the image-text pairs from CODA-LM into a dialogue format and employ a rational data sampling strategy to form an instruction-following dataset. 
Furthermore, to efficiently learn while preserving as much pre-training knowledge as possible, we use LoRA~\cite{hu2021lora} to fine-tune both the LLM and the visual encoder\footnote{https://huggingface.co/openai/clip-vit-large-patch14-336}.

\paragraph{Visual representation.}
To obtain more effective visual representations and enhance the model's regional perception capabilities, we refer to the dynamic high resolution (\ie, AnyRes) from LLaVA-NeXT~\cite{liu2024llavanext}. 
While retaining the fixed global image resolution, we split original images into different sub-images, each independently encoded by a shared visual encoder, and finally concatenate all visual tokens together before feeding into LLMs. 
Moreover, considering context lengths and training costs of LLMs, we observe that a 2$\times$2 MaxPool operation on visual tokens of sub-images can effectively reduce redundancy, achieving a better trade-off between efficiency and performance.

\paragraph{Implementation details.}
It is worth noting that our approach is simple yet effective. The training of CODA-VLM requires only ~3 hours on 8 A800 GPUs. Specifically, we use LoRA with r = 256 and $\alpha$ = 256 for the LLM, and r = 64 and $\alpha$ = 16 for the visual encoder, fine-tuning with a context length of 4096. The learning rate is set to $2e^{-4}$, training for 4 epochs with a batch size of 16 per GPU. 
We utilize the combination of the train and validation splits of CODA-LM, as discussed in Sec.~\ref{sec:collection}.
In Sec.~\ref{sec:analysis} and~\ref{sec:ablate_codavlm}, we provide more detailed analysis and empirical ablation results on CODA-VLM.

%% file: sec/5_Evaluation_Results.tex

\begin{table*}[t]
    \centering
    \setlength{\tabcolsep}{1.5mm}
    \scalebox{1}{
    \begin{tabular}{l|c|c|ccccccc|c}
        \toprule
        \multirow{2}{*}{\textbf{Method}} & {\textbf{General$\uparrow$}} & \multicolumn{8}{c|}{
        \textbf{Regional Perception $\uparrow$}} & \textbf{Suggestion$\uparrow$} \\
        \cline{3-10}
        & \textbf{Text-Score} & \textbf{ALL} & \textbf{Vehicle} & \textbf{VRU} & \textbf{Sign} & \textbf{Light}  & \textbf{Cone} & \textbf{Barrier} & \textbf{Other} & \textbf{Text-Score} \\
        \midrule
        MiniGPT-v2-7B & 11.58 & 15.93 & 18.74 & 13.58 & 15.71 & 17.78 & 15.34 & 13.02 & 14.41 & 10.00 \\
        Shikra-7B & 12.24 & 22.94 & 28.29 & 17.88 & 20.00 & 15.56 & 21.23 & 20.00 & 19.67 & 10.20 \\
		LLaVA1.5-7B & 19.30 & 42.06 & 46.67 & 38.47 & 39.14 & 48.89 & 50.83 & 30.93 & 33.82 & 23.16 \\
        Qwen-VL-Chat-7B & 18.22 & 26.62 & 35.48 & 24.16 & 20.86 & 23.33 & 19.61 & 17.56 & 25.86 & 22.06 \\
        MiniCPM-V-2.5-8B & 41.12 & 57.20 & 61.91 & \underline{54.82} & \underline{59.43} & 46.67 & 66.57 & 35.35 & \underline{58.75} & 48.48 \\
        \midrule
        LLaVA1.5-13B & 24.54 & 42.41 & 53.62 & 36.79 & 33.71 & 46.67 & 41.27 & 30.41 & 33.82 & 27.90 \\
        LLaVA-NeXT-13B & 29.86 & 53.63 & 55.51 & 47.08 & 54.00 & \underline{60.00} & 70.34 & 40.47 & 46.45 & 31.92 \\
        InternVL-V1-5-20B & 38.38 & \underline{61.53} & \underline{63.77} & 53.14 & 50.57 & 57.78 & \underline{80.34} & 46.86 & 57.11 & 41.18 \\
        \midrule
        Gemini-Pro & 25.24 & 51.38 & 49.03 & 42.77 & 37.43 & 42.22 & 69.56 & 45.70 & 51.32 & 27.40 \\
        GPT-4V & \textbf{57.50}  & 56.26 & 60.89 & 40.58 & 49.43 & 54.44 & 66.08 & \underline{50.17} & 53.16 & \textbf{63.30} \\
        \midrule
        \rowcolor{backcolor}
        \textbf{CODA-VLM (ours)} & \underline{55.04} & \textbf{77.68} & \textbf{78.79} & \textbf{73.80} & \textbf{64.86} & \textbf{73.33} & \textbf{86.18} & \textbf{78.72} & \textbf{68.75} & \underline{58.14} \\
        \bottomrule
    \end{tabular}
    }
    \caption{\textbf{Comparison among open-sourced and commercial LVLMs on CODA-LM Test set.}
    All open-sourced LVLMs suffer from the complicated road corner cases, while our \textbf{CODA-VLM}, due to its usage of superior vision representation and knowledge transfer, performs the best or second best on all evaluated dimensions, surpassing all open-sourced counterparts.
    Note that here we re-scale the original 1-10 \texttt{Text-Score} to 1-100 for better readability.
    \textbf{Bold} denotes the best results, while \underline{underline} suggests the second best.
    }
    \label{tab:comparison}
\end{table*}


\section{CODA-LM Benchmark}
In this section, based on the proposed CODA-LM dataset, we start by comparing and analyzing the performance of different LVLMs in Sec.~\ref{sec:result}, followed by an in-depth analysis of model architecture designs in Sec.~\ref{sec:analysis}.
We then conduct an ablation study on critical components of dataset construction and evaluation in Sec.~\ref{sec:abl}.


\subsection{Main Results}\label{sec:result}

\paragraph{Baselines.}
In this work, we evaluate a total of 10 LVLMs, including both open-sourced and commercial models. 
Commercial models consist of the Gemini-Pro~\cite{gemini} and GPT-4V~\cite{gpt4v}, while the open-sourced LVLMs are categorized based on the parameter sizes of their language models. 
The 7B/8B variants include the MiniGPT-v2~\cite{chen2023minigpt}, Shikra~\cite{chen2023shikra}, LLaVA1.5~\cite{liu2023improved}, Qwen-VL-Chat~\cite{bai2023qwen} and MiniCPM-Llama3-V-2.5~\cite{MiniCPM-V}, while the 13B/20B LVLMs consist of LLaVA1.5~\cite{liu2023improved}, LLaVA-NeXT~\cite{liu2024llavanext} and InternVL-Chat-V1-5 \cite{chen2024far}.
Each model is evaluated on the three tasks separately for a comprehensive analysis of their performance on self-driving corner cases. 

\begin{figure*}[t]
  \centering
\includegraphics[width=1.0\linewidth]{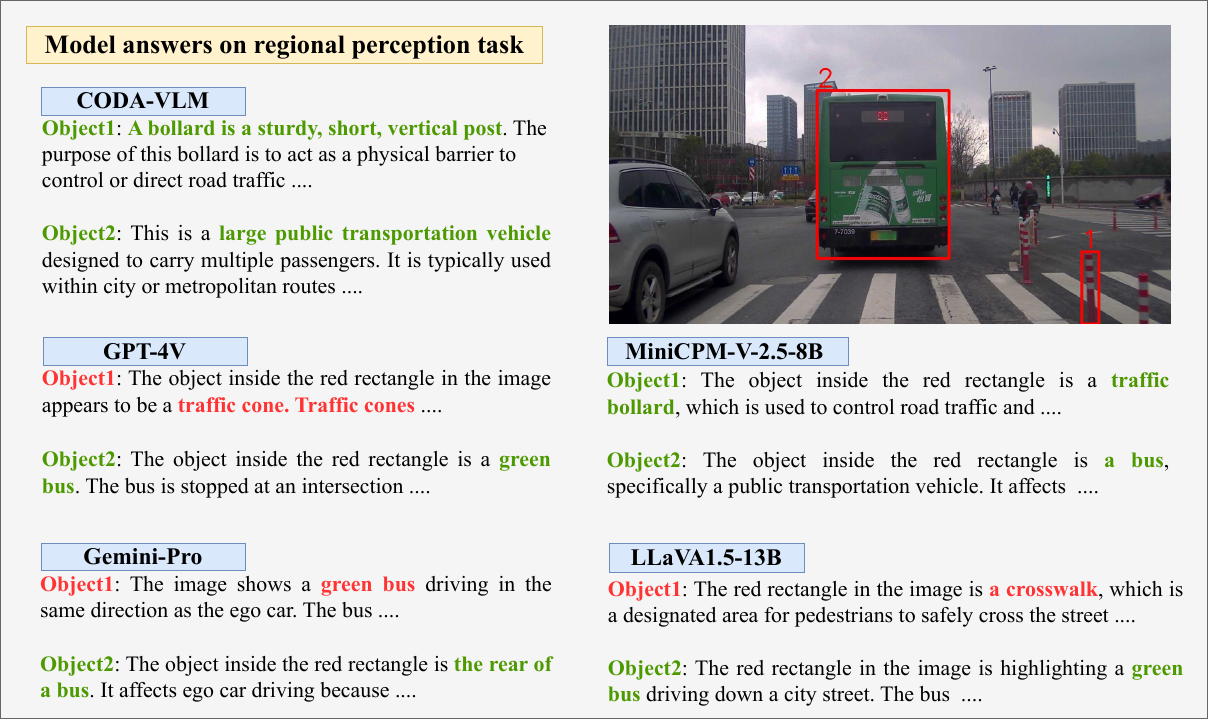}
  \caption{
  \textbf{
  Qualitative comparison among different LVLMs on the regional perception task. }Mistakes within the model responce are highlighted in \textbf{\textcolor[RGB]{255,51,51}{red}}, whereas the accurate parts are emphasized in \textbf{\textcolor[RGB]{77,153,0}{green}}.
  }
  \label{fig:TvsV}
\end{figure*}

\paragraph{Setting.}
To ensure the reproducibility of our evaluation results, we use the same prompt for generating responses for all evaluated LVLMs and employ greedy decoding during inference, which generates the next token with the highest probability at each step as output, thus eliminating randomness during inference. 
As discussed in Sec.~\ref{sec:evaluation}, GPT-4 is used as the judge for evaluation, with the temperature coefficient set to 0 and a fixed random seed, to ensure consistency when scoring different models.
\paragraph{Results.}
The comparison results on the CODA-LM Test set are reported in Tab.~\ref{tab:comparison}. 
Among the open-sourced baselines, MiniCPM-V-2.5-8B achieves the best performance, probably due to the usage of the powerful LLaMA3 base model, only ranking second to Intern-VL-1.5-20B on regional perception.
Among the commercial models, GPT-4V continues to demonstrate a leadership position, ranking first on general perception and driving suggestions.
Interestingly, Gemini-Pro is polarized, showing poor results in general perception and driving suggestions while excelling in regional perception. 
\textbf{CODA-VLM}, instead, achieves the best or second best on all the evaluated dimensions, surpassing all open-sourced counterparts.
CODA-VLM obtains comparable performance with GPT-4V, even exceeding GPT-4V by \textbf{+21.42\%} on regional perception.
A qualitative comparison is given in Fig.~\ref{fig:TvsV}.

\subsection{Analysis}\label{sec:analysis}
\paragraph{Visual represention.}
Recent works~\cite{liu2024llavanext,chen2024far} have revealed the significant benefit of utilizing high-resolution images as input for LVLMs.
For regional perception, simply increasing the image resolution from 224 to 336 enables LLaVA1.5-7B to outperform Shikra-7B by 20\%.
By further increasing the effective resolution with the AnyRes, LLaVA-NeXT-13B surpasses the LLaVA1.5-13B by over 11\%. 
The compression of visual tokens is another factor.
Even with a 448 image resolution, Qwen-VL-Chat-7B is 16\% lower than LLaVA1.5-7B with 336 image inputs, largely due to the usage of Q-former for token compression.
In contrast, InternVL-V1-5-20B merges four adjacent tokens, while MiniCPM-LLaMA3-V-2.5 resamples each sub-image individually, both effectively reducing redundant tokens while maximizing performance retention. 
The same tendency can be observed in general perception and driving suggestions tasks.
Therefore, in CODA-VLM, we adopt AnyRes with a 2$\times$2 MaxPool to achieve the balance between performance and efficiency.

\paragraph{Knowledge transfer.}
The knowledge embedded in LVLMs significantly influences the performance, which, on the one hand, comes from pre-trained visual encoders and LLMs, while on the other hand, also arises from high-quality visual instruction fine-tuning. 
As reported in Tab.~\ref{tab:comparison}, MiniCPM-V-2.5-8B surpasses LLaVA-NeXT-13B by 12\% and 17\% in general perception and driving suggestions, despite having smaller LLMs, revealing the significance of LLaMA3-8B.
Moreover, we observe that GPT-4V exceeds open-sourced LVLMs by a significant margin on general perception and driving suggestions, indicating that current open-sourced LVLMs still lack the domain-specific knowledge of self-driving.
Therefore, in CODA-VLM, we adopt LLaMA3-8B as our base model and conduct the domain-specific fine-tuning with driving scenes in CODA-LM.


\begin{table}[t]
    \centering
    \setlength{\tabcolsep}{1.5mm}
    \begin{tabular}{l|c|ccc}
        \toprule
        Judge & Reference & General & Regional & Suggestion \\
        \midrule
        GPT-4 & GT & \cellcolor{baselinecolor}\textbf{83.67} & \cellcolor{baselinecolor}\textbf{85.71} & \cellcolor{baselinecolor}\textbf{89.80} \\ 
        GPT-4V & Image & 69.39 & 75.51 & 69.39 \\
        GPT-4V & Img \& GT & 79.59 & 79.59 & 87.76 \\
        \bottomrule
    \end{tabular}
    \caption{
    \textbf{Consistency between different judges and human judgments.}
    Text-only GPT-4 judges reveal superior consistency for all tasks.
    GT denotes ground truth answers.
    Default settings are marked in \colorbox{baselinecolor}{gray}.
    }
    \label{tab:human consis}
\end{table}


\begin{table}[t]
    \centering
    \setlength{\tabcolsep}{1.5mm}
    \begin{tabular}{l|c|c}
        \toprule
        Judge & Reference & Consistency (\%) \\
        \midrule
        GPT-4 & Plain & 71.43 \\
        GPT-4 & Structured \& Concat & 77.55 \\
        GPT-4 & Structured \& Coherent & \cellcolor{baselinecolor}\textbf{83.67} \\
        \bottomrule
    \end{tabular}
    \caption{
    \textbf{Consistency among human judgments and GPT-4 judges with different references.}
    The \textit{structured coherence} manner reveals significant superiority.
    }
    \label{tab:human consis reference}
\end{table}


\subsection{Ablation Study}\label{sec:abl}


\subsubsection{Human Consistency of Judges}\label{sec:consistency}
Following LMSYS~\cite{zheng2024judging}, we adopt the ranking-based manner to calculate the consistency of the GPT-4 and GPT-4V judges with human judgments.
We randomly sample 50 samples from the CODA-LM Test set, and for each sample, we further sample two model responses from Tab.~\ref{tab:comparison}, followed by random shuffling.
We then ask judges to determine the ranking (with ties) of the two candidate responses and human consistency is calculated as the probability of the GPT judge agreeing on the ranking with human judgments.

As reported in Tab.~\ref{tab:human consis}, the text-only GPT-4 judge with the reference answers achieves more than 80\% consistency for all three tasks, surpassing the GPT-4V variants by a large margin.
The GPT-4V judge suffers when only images are provided as the reference, which is relieved when reference answers are provided, but still inferior to the text-only GPT-4 judge, even with a higher expense.


\begin{table}[t]
    \centering
    \begin{tabular}{l|cc}
        \toprule
        Method & Grounding & Visualization \\
        \midrule
        Shikra-7B & 20.39 & \textbf{22.94}\small\textcolor{improvecolor}{$^{+2.55}$} \\
        LLaVA1.5-13B & 18.41 & \textbf{42.41}\small\textcolor{improvecolor}{$^{+24.0}$} \\
        GPT-4V & 12.85 & \textbf{56.26}\small\textcolor{improvecolor}{$^{+43.41}$} \\
        \bottomrule
    \end{tabular}
    \caption{
    \textbf{Ablation on visual prompts for regional perception.}
    Visualization with red rectangle boxes shows consistent improvements among all evaluated models.
    }
    \label{tab:ablation_region_data}
\end{table}


\begin{table}[t]
    \centering
    \setlength{\tabcolsep}{1.5mm}
    \scalebox{0.94}{
    \begin{tabular}{l|c|c|c}
        \toprule
        \multirow{2}{*}{Model} & Training & General & Driving \\
        & Time & Perception & Suggestion \\
        \midrule
        \textbf{LLaVA-1.5} & - & 15.84 & 29.24 \\
        + Drive SFT Data & 1.5h & 53.35 & 60.83 \\
        + CLIP LoRA & 1.6h & 53.65 & 61.17 \\
        + AnyRes & 6h & \textbf{57.46} & \textbf{61.83} \\
        + 2$\times$2 MaxPool & 3h & \cellcolor{baselinecolor}56.04 & \cellcolor{baselinecolor}61.42 \\
        \bottomrule
    \end{tabular}
    }
    \caption{\textbf{Ablation on our CODA-VLM components.}
    Training time (hours) is estimated with 8$\times$ A800 GPUs.
    }
    \label{tab:model}
\end{table}


\subsubsection{Hierarchical Data Structure for General Perception}

We ablate the necessity of using the ``structured-coherence'' pipeline in Tab.~\ref{tab:human consis reference}.
Following Sec.~\ref{sec:consistency}, we evaluate the quality of pre-annotations by using them as the reference for the GPT-4 judge and then calculate the consistency with human judgments.
We compare with 1) plain text prompting and 2) structured prompting followed by concatenating annotations of each category to consecutive texts.
As shown in Tab.~\ref{tab:human consis reference}, generating structured responses followed by coherence obtains the best consistency.


\subsubsection{Visual Prompts for Regional Perception}

We ablate the advantage of using visualization over grounding as visual prompts for regional perception.
The prompt for visualization is ``\textit{Please describe the object inside the red rectangle in the image and explain why it affects ego car driving}'', while the prompt for grounding is ``\textit{Please provide a description for this object and explain why this object affects ego car driving: [x1, y1, x2, y2]}''. 
As reported in Tab.~\ref{tab:ablation_region_data}, visualization demonstrates consistent improvement for all evaluated LVLMs, even for Shikra-7B which has been pre-trained with grounding data specifically. 

\subsubsection{CODA-VLM Components}\label{sec:ablate_codavlm}
We ablate the usage of different components of CODA-VLM on a 200-image subset of the CODA-LM Test set.
Starting from a pre-trained LLaMA3-8B-based LLaVA1.5 checkpoint, we ablate the usage of 1) domain-specific fine-tuning, 2) training CLIP encoder with LoRA, 3) adopting AnyRes and 4) conducting 2$\times$2 MaxPool step by step.
As shown in Tab.~\ref{tab:model}, our CODA-VLM achieves a better trade-off among efficiency and performance.

\subsection{Limitations}

CODA-LM is built on corner cases from CODA, which might not cover all possible unexpected conditions in driving scenarios, and we opt to explore controllable generation~\cite{chen2023integrating,gao2023magicdrive,gao2024magicdrive3d,li2023trackdiffusion,liu2023geomerasing,wang2024detdiffusion} to generate corner cases in the future.
CODA-LM focuses on interpretable self-driving, and we will explore collecting action-level annotations.
The current data collection pipeline relies on human verification and revision to ensure the quality of annotations, and an automatic data calibration method is also appealing.
How to better incorporate visual pre-trained prior (\eg, self-supervised learning~\cite{chen2021multisiam,chen2023mixed,liu2022task,zhili2023task}) is also open.

%% file: sec/6_conclusion.tex
\section{Conclusion}
In this paper, we propose CODA-LM, a novel real-world multimodality road corner case dataset for autonomous driving with a hierarchy task framework, spanning from general and regional perception to driving suggestions, to support automated evaluation of Large Vision-language Models (LVLMs) on self-driving corner cases.
We conduct a comprehensive evaluation of representative LVLMs on road corner cases and propose CODA-VLM, a novel driving LVLM specialized in driving perception and suggestions.
However, we are still far from a fully intelligent driving agent and we hope our CODA-LM can serve as the catalyst to promote the development of reliable and interpretable autonomous driving systems.

%% file: sec/7_appendix.tex

 \begin{figure}[t]
  \centering
\includegraphics[width=\linewidth]{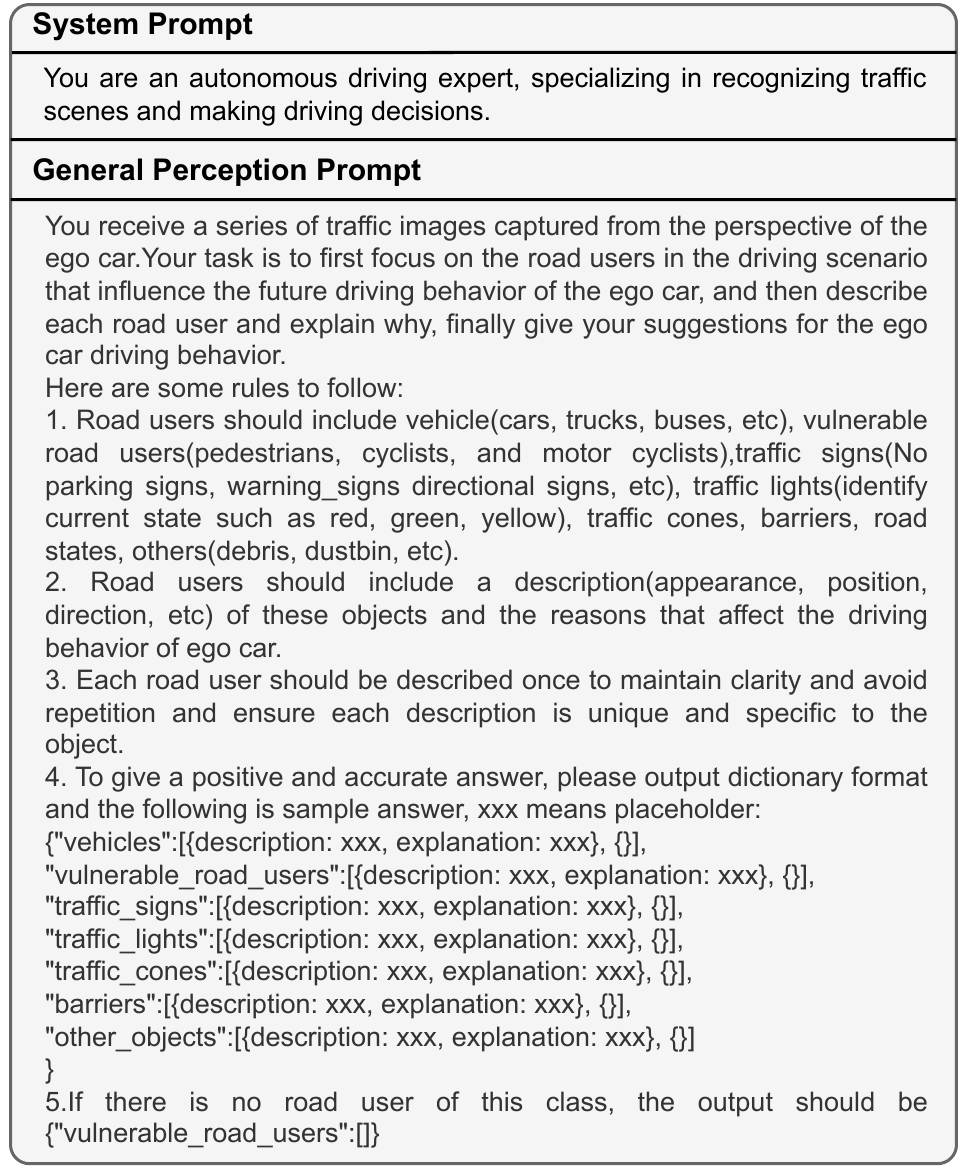}
  \vspace{-8mm}
  \caption{
  \textbf{The data pre-annotation prompts for general perception.}
  The prompts are divided into system prompts and general perception prompts.
  }
  \label{fig:gen_pr}
\end{figure}

 \begin{figure}[t]
  \centering
\includegraphics[width=\linewidth]{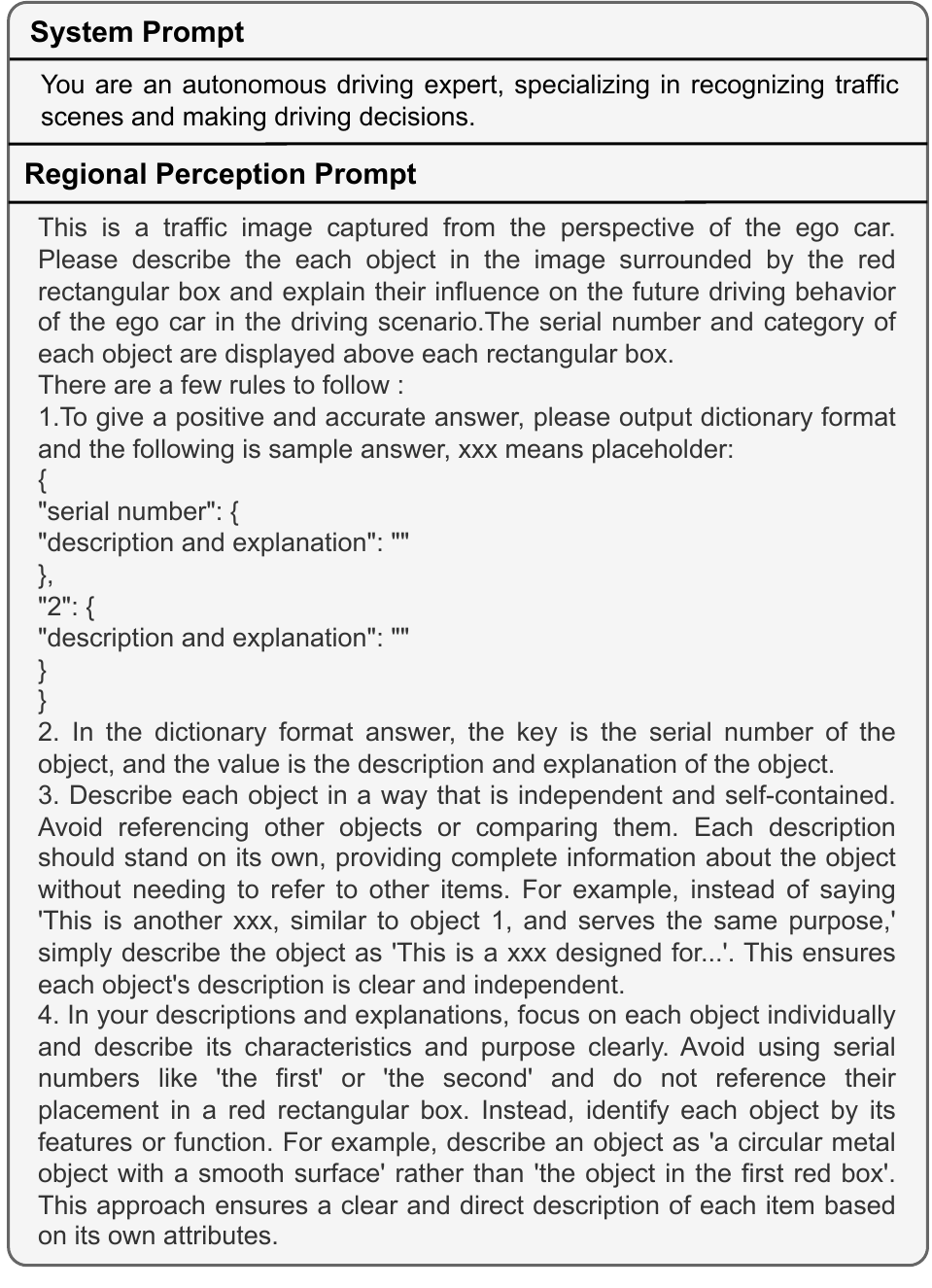}
  \vspace{-8mm}
  \caption{
  \textbf{The data pre-annotation prompts for regional perception.}
  The prompts are divided into system prompts and regional perception prompts.
  }
  \label{fig:reg_pr}
\end{figure}


\appendix
\section*{Appendix}
\section{More on Dataset Construction}\label{app_prompt}

\paragraph{Prompts for pre-annotation.}
The prompts used to generate the pre-annotations from GPT-4V are provided in Fig.~\ref{fig:gen_pr} and Fig.~\ref{fig:reg_pr}.

\begin{figure*}[t]
  \centering
\includegraphics[width=\linewidth]{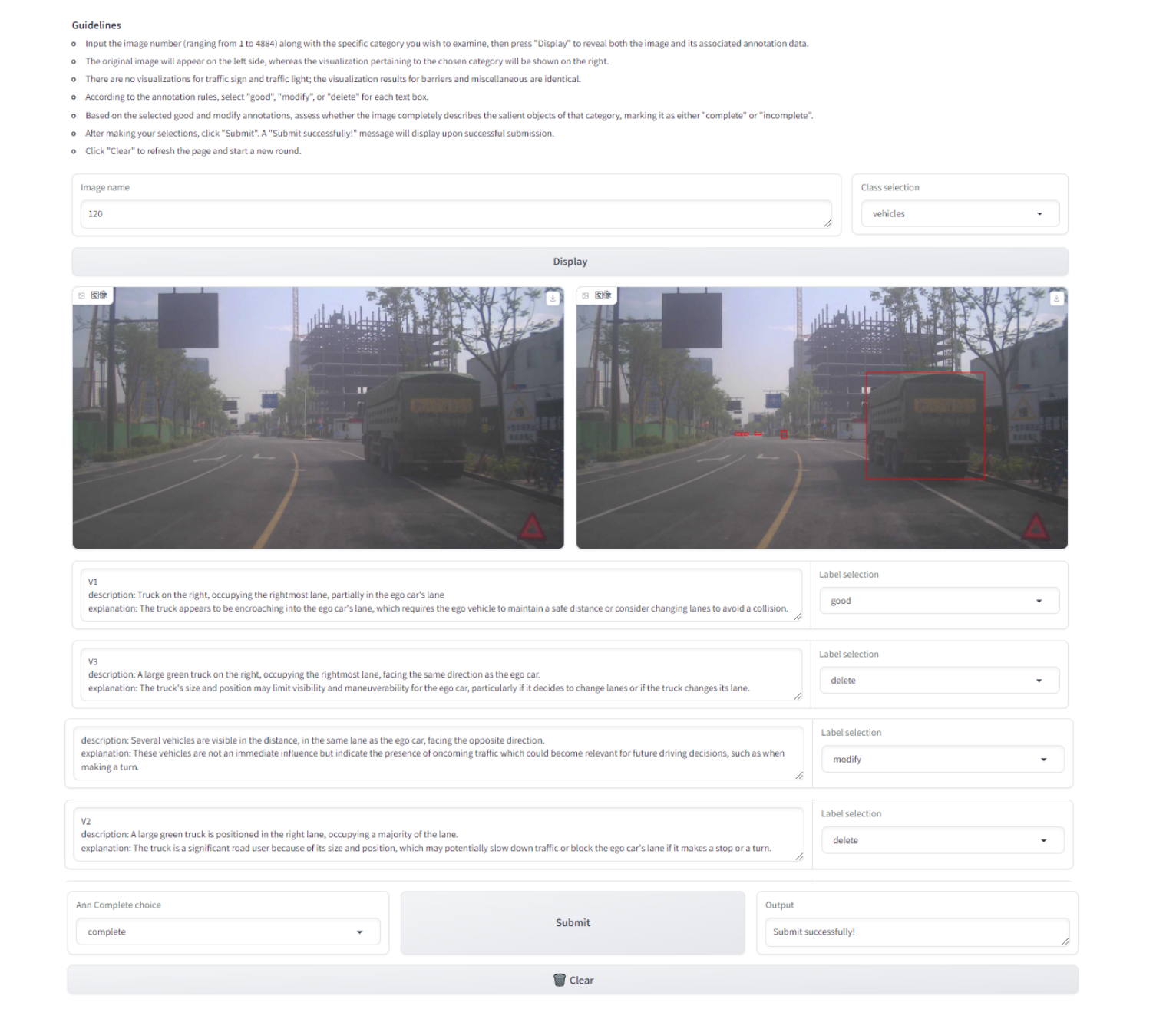}
  \caption{
  \textbf{Our Gradio labeling tool graphics user interface} for general perception.
  }
  \label{fig:label_tool}
\end{figure*}

\paragraph{Gradio labeling tool graphical user interface (GUI).}
Fig.~\ref{fig:label_tool} demonstrates a screenshot of our labeling tool for the general perception task. We utilize Gradio and aim to assist human annotators to refine general perception pre-annotations deriving from GPT-4V, as discussed in Sec.~\ref{sec:collection}. 
The annotators refine by following the principles of merging, modifying, and deleting step by step.

\begin{figure}[t]
  \centering
\includegraphics[width=1.0\linewidth]{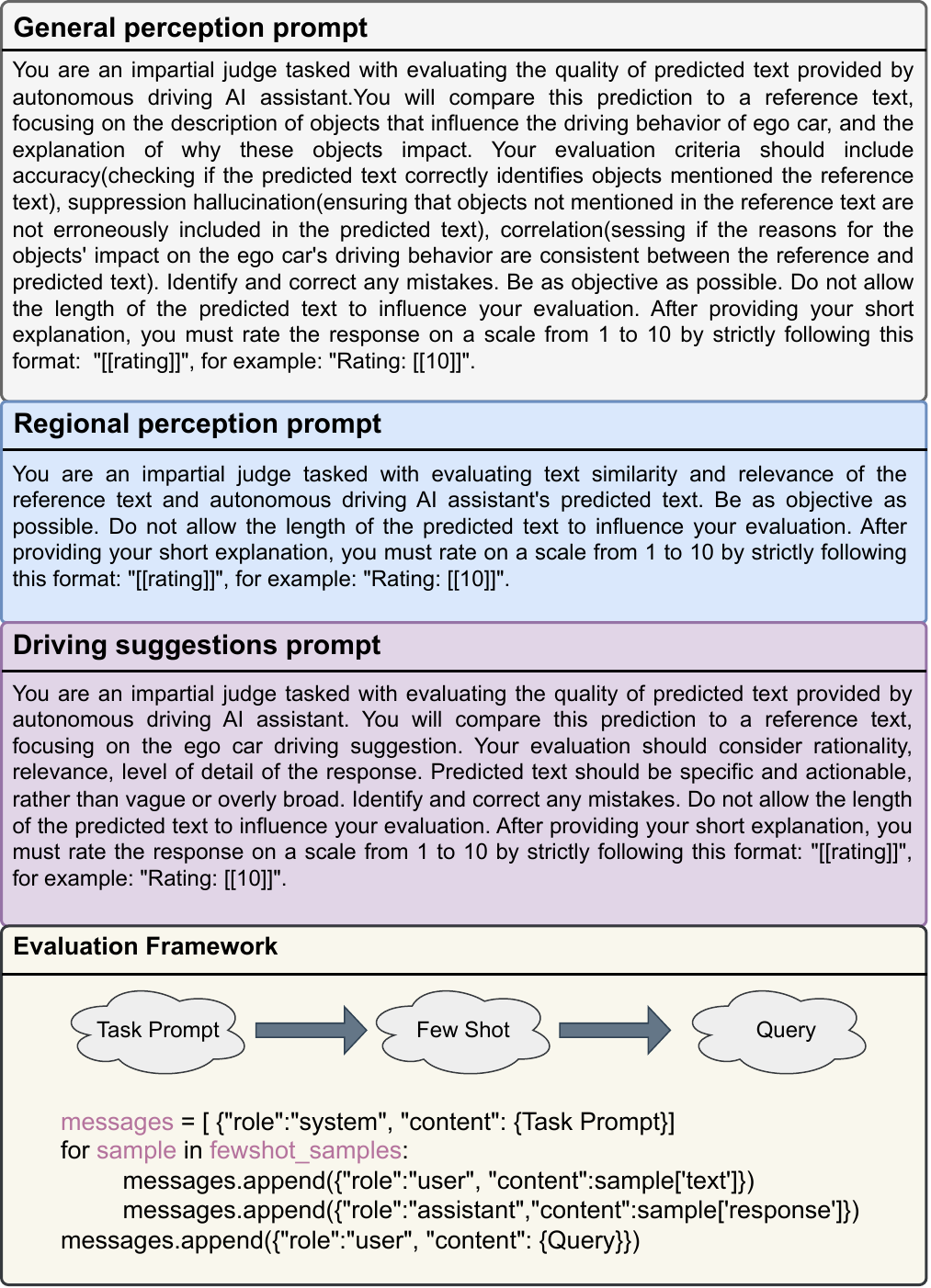}
  \vspace{-8mm}
  \caption{
  \textbf{Evaluation framework of CODA-LM.} 
    We utilize text-only GPT-4 judges empowered by ICL few-shot examples to evaluate LVLMs on CODA-LM.
  }
  \label{fig:eva_pr}
\end{figure}

\begin{figure}[t]
  \centering
\includegraphics[width=1.0\linewidth]{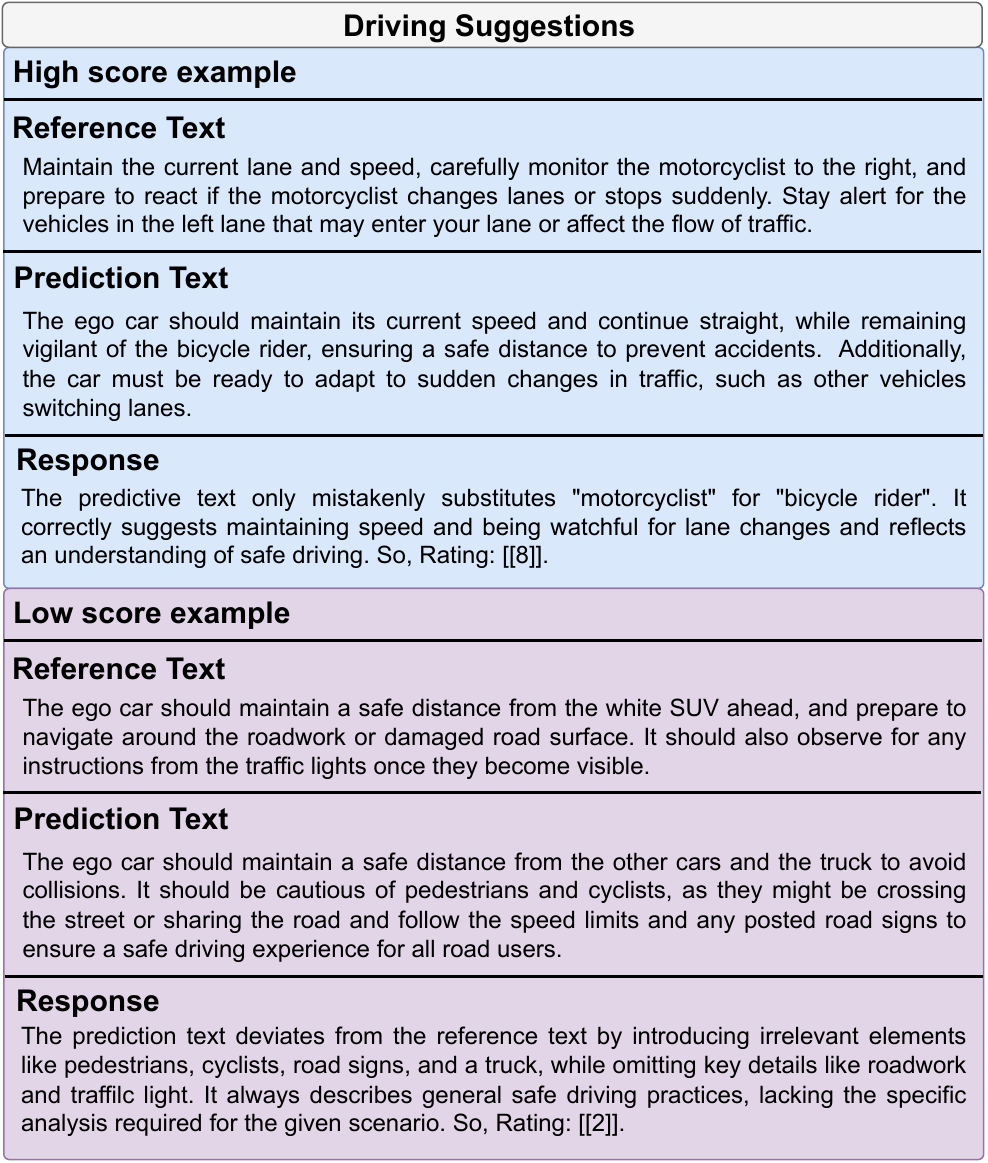}
\vspace{-8mm}
  \caption{
  \textbf{Few-shot examples for Driving Suggestions}
  }
  \label{fig:driving_suggestions_example}
\end{figure}

\begin{figure}[t]
  \centering
\includegraphics[width=1.0\linewidth]{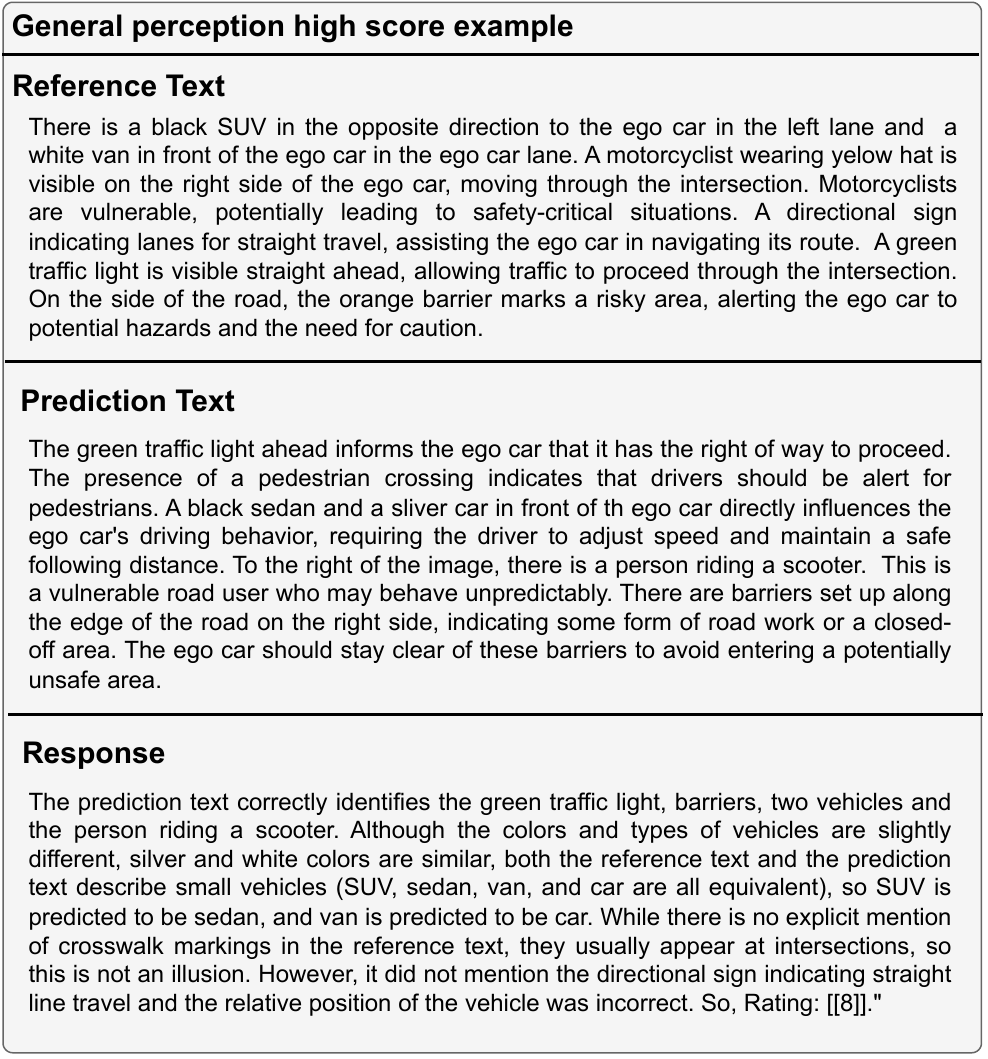}
  \caption{
  \textbf{The high score few-shot example for general perception.}
  }
  \label{fig:high_few_shot}
\end{figure}

\begin{figure}[t]
  \centering
\includegraphics[width=1.0\linewidth]{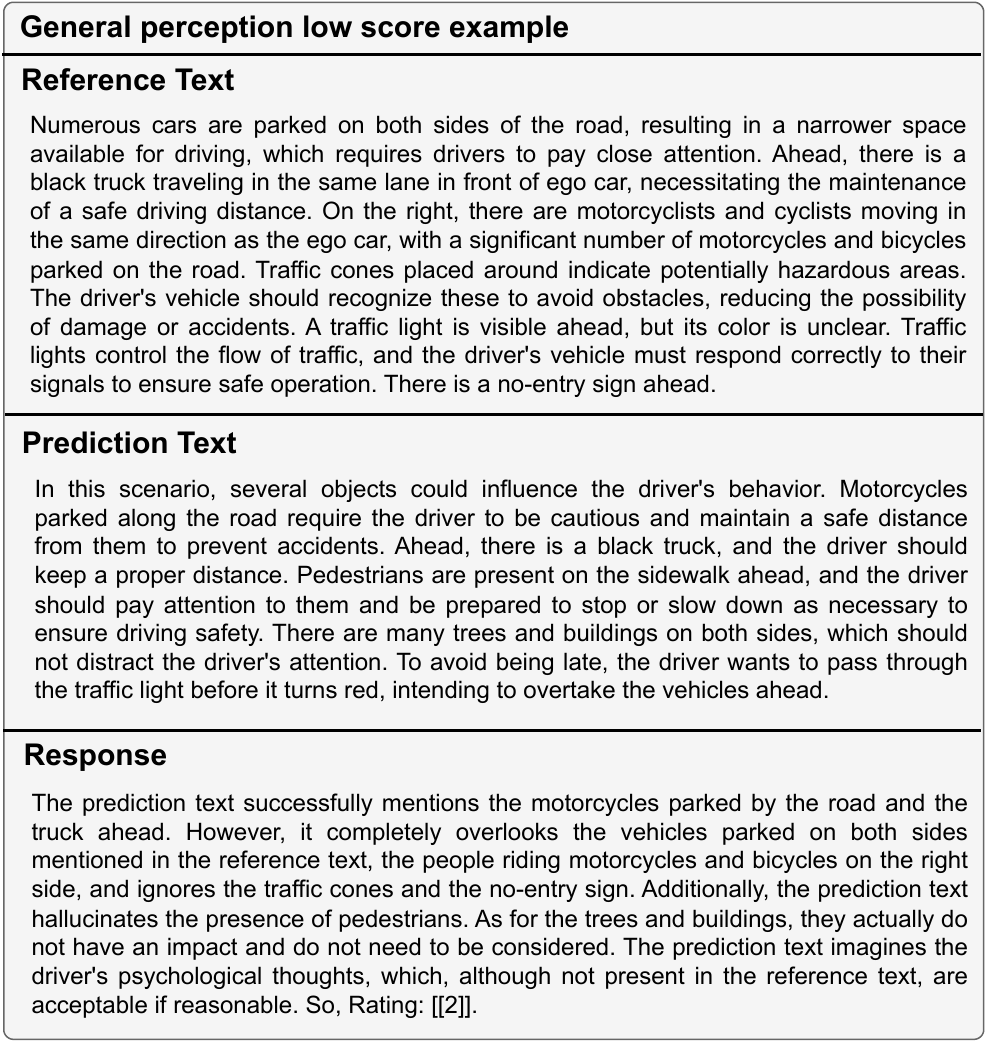}
  \caption{
  \textbf{The low score few-shot example for general perception.}
  }
  \label{fig:low_few_shot}
\end{figure}

\paragraph{Prompts for evaluation.}
To comprehensively and accurately assess the performance of different LVLMs, we design distinct evaluation prompts for each task, as shown in Fig.~\ref{fig:eva_pr}. Meanwhile, we use the few-shot in-context learning method to improve accuracy for general perception and driving suggestions. Specifically, we design in-context examples with different scores to assist judgement. Please see few-shot in-context-learning examples for general perception in Fig.~\ref{fig:high_few_shot} and Fig.~\ref{fig:low_few_shot} for details. Additionally, few-shot in-context-learning examples for driving suggestions are in Fig.~\ref{fig:driving_suggestions_example}. 


\section{More Experiments}\label{app:exp}

\paragraph{Evaluation metrics.}
When conducting a corner case regional perception evaluation, the data is organized in the form of brief sentences. Therefore, in addition to using the Text-Score for evaluation, we also explore the impact of traditional keyword-based metrics, including BLEU-4~\cite{papineni2002bleu}, METEOR~\cite{banerjee2005meteor}, CIDEr~\cite{vedantam2015cider}, and SPICE~\cite{anderson2016spice}, as shown in Tab.~\ref{tab:ablation_suggest}. 
For better demonstration, we multiplie the scores by 100, normalizing them to a range of 1-100, similarly with the Text-Score. BLEU-4 primarily evaluates quality through lexical matching and cannot capture the semantic accuracy of the generated text. CIDEr is not suitable for texts with low lexical repetition. Hence, the scores from these two metrics do not reflect performance accurately. Although METEOR can account for synonyms, it still does not reflect the actual semantics, so despite some differences in scores, they are not accurate. In contrast, SPICE can reflect semantic accuracy to some text, and even though the overall scores are still low, it successfully indicates the trend among different models, with InternLM2-vl still leading among open-source models.
By default, we still adopt the Text-Score as the primary evaluation metric, unless otherwise specified.


\section{Qualitative Comparison}\label{answer_diff}
\begin{figure*}[htb]
  \centering
\includegraphics[width=1.0\linewidth]{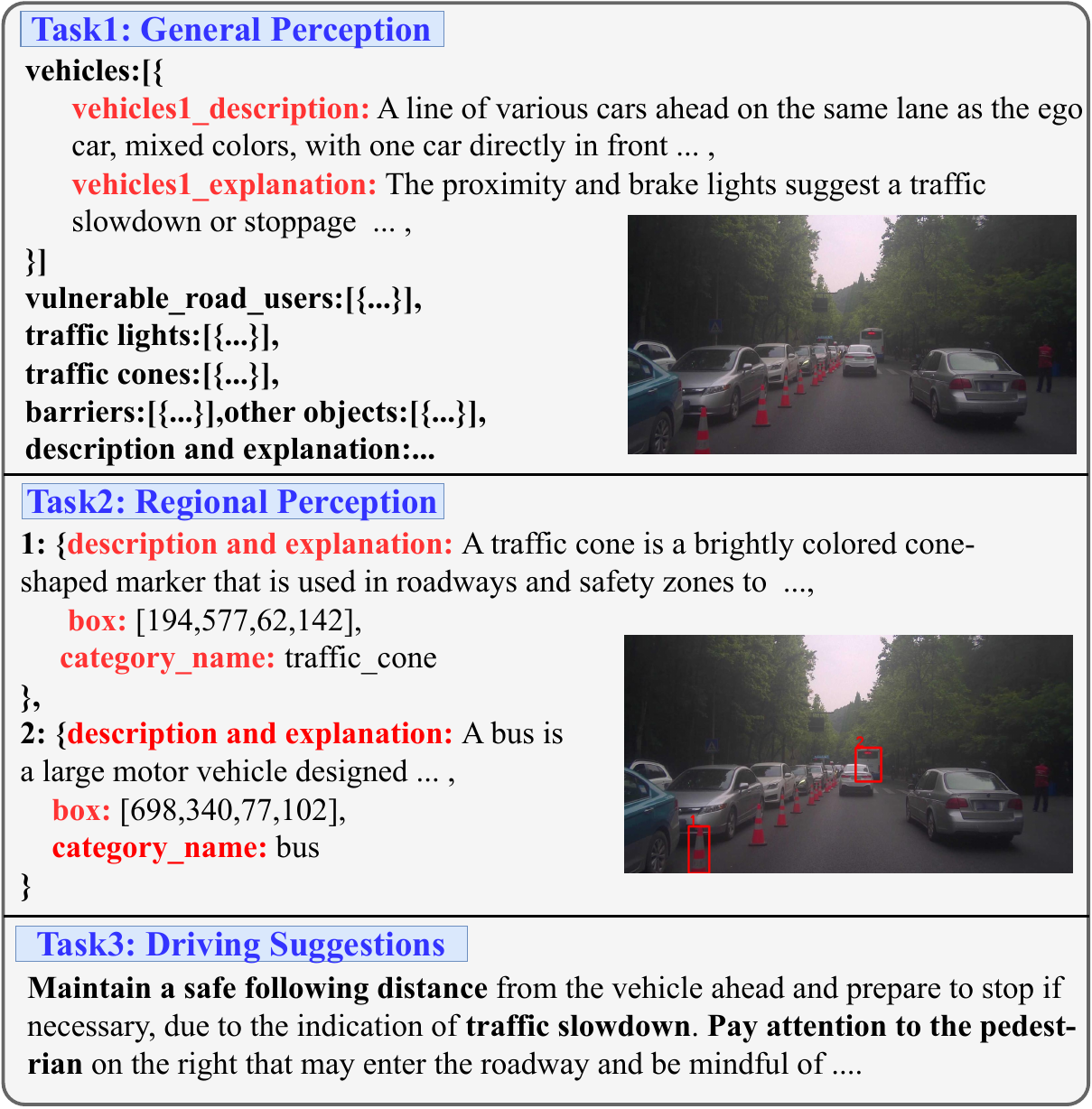}
  \caption{
  \textbf{
  More data examples of CODA-LM.}
  }
  \label{fig:ov1}
\end{figure*}

\begin{figure*}[htb]
  \centering
\includegraphics[width=1.0\linewidth]{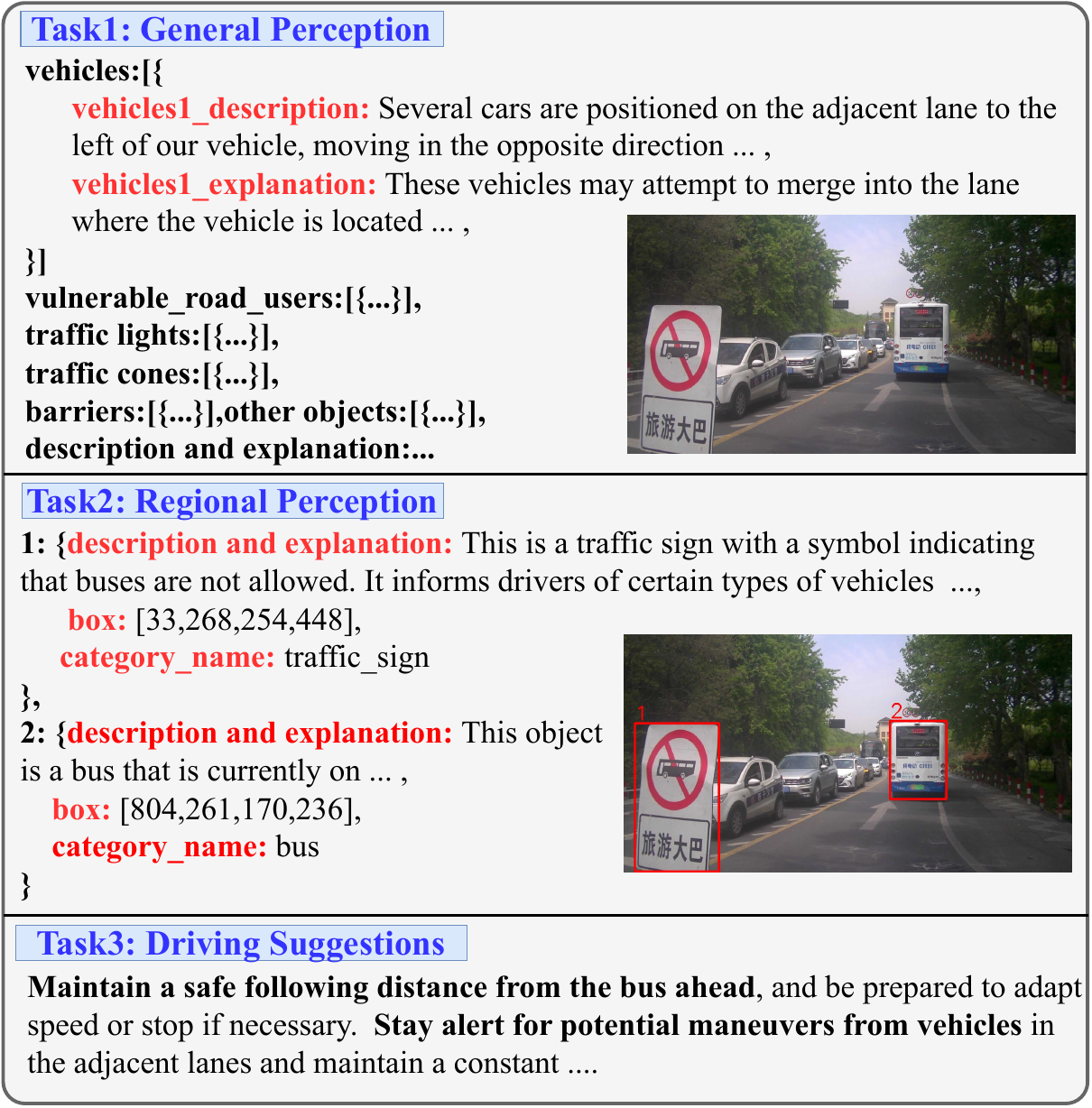}
  \caption{
  \textbf{
  More data examples of CODA-LM.}
  }
  \label{fig:ov2}
\end{figure*}

\begin{figure*}[htb]
  \centering
\includegraphics[width=1.0\linewidth]{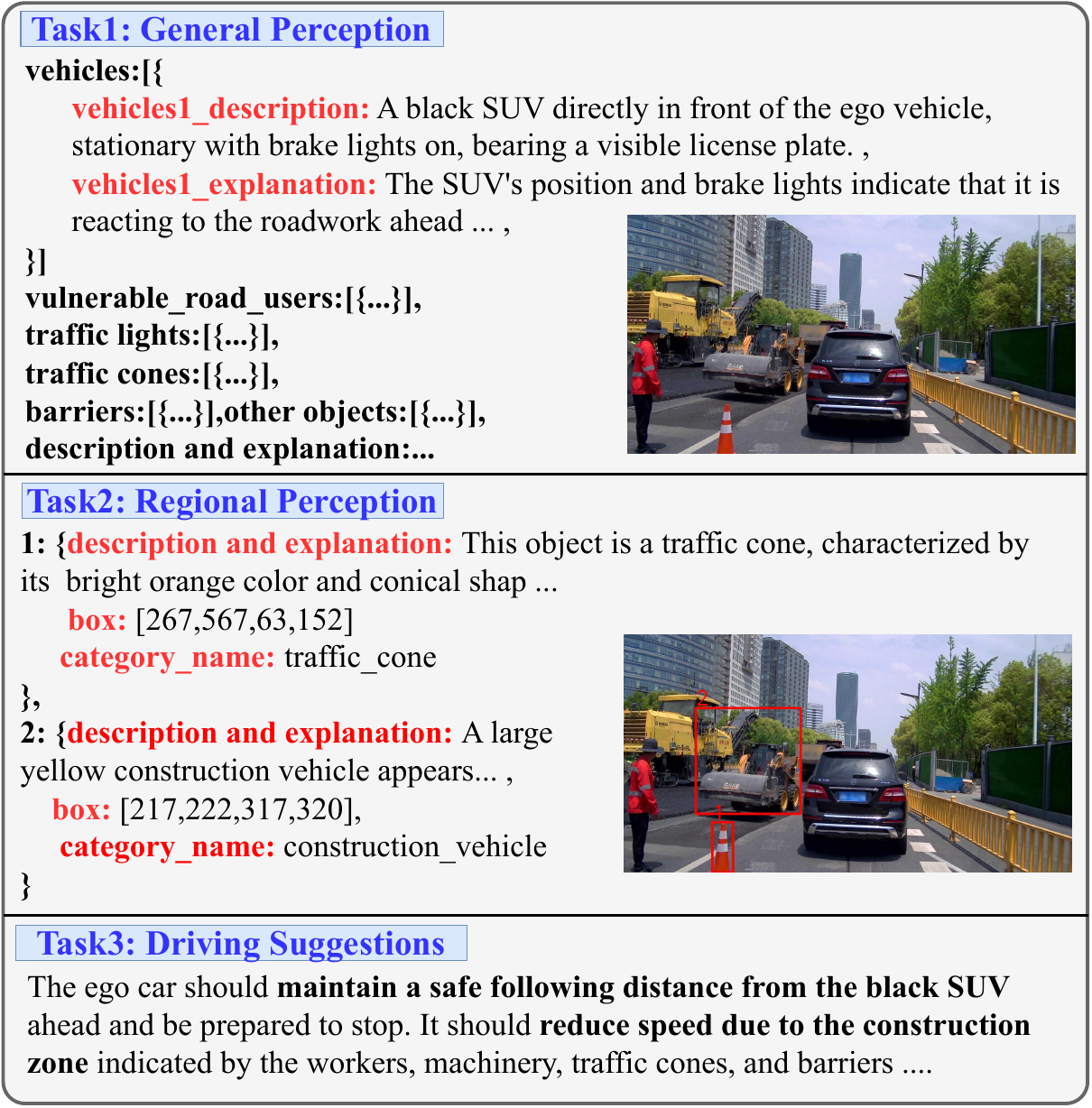}
  \caption{
  \textbf{
  More data examples of CODA-LM.}
  }
  \label{fig:ov3}
\end{figure*}

\begin{figure*}[htb]
  \centering
\includegraphics[width=1.0\linewidth]{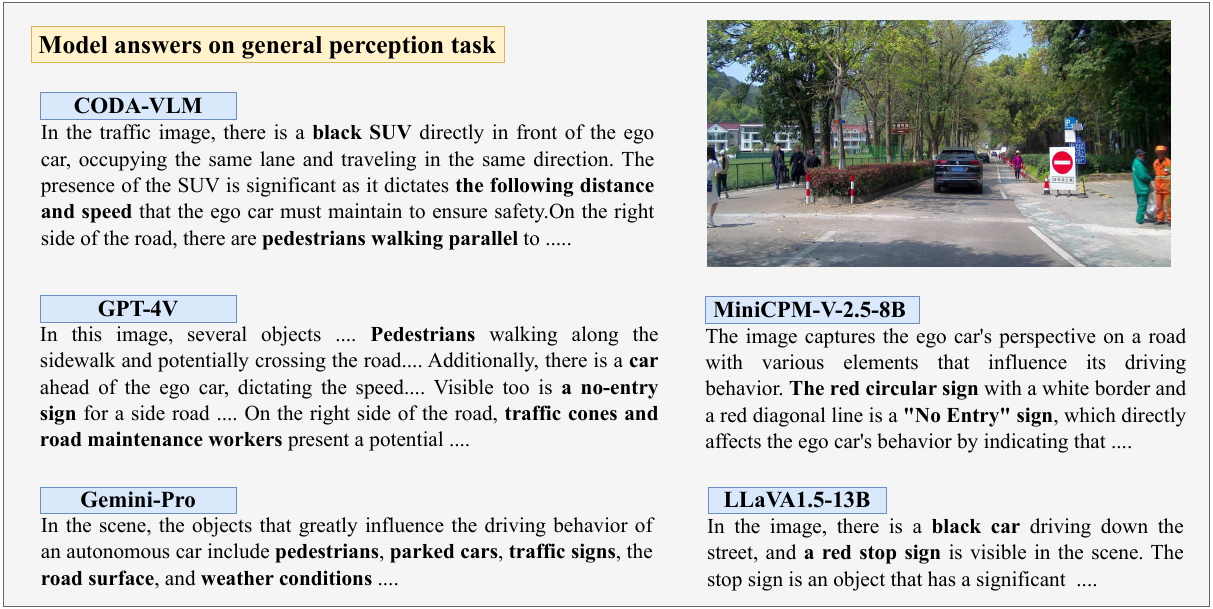}
  \caption{
  \textbf{
  More qualitative comparison for the general perception task. }
  }
  \label{fig:re1st1}
\end{figure*}

\begin{figure*}[htb]
  \centering
\includegraphics[width=1.0\linewidth]{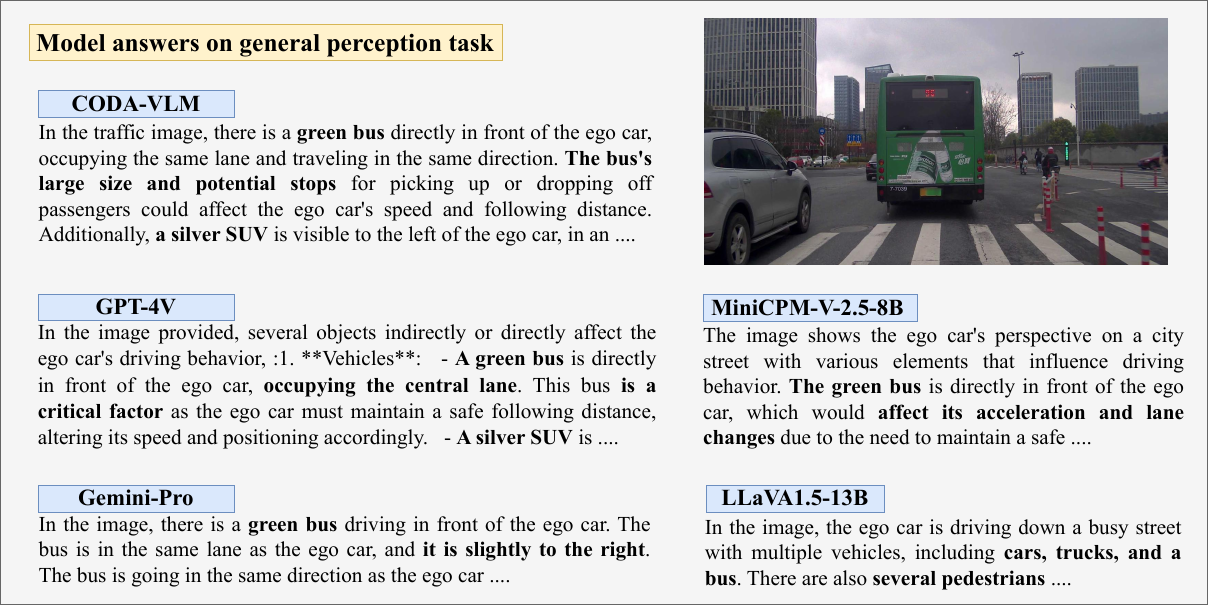}
  \caption{
  \textbf{
  More qualitative comparison for the general perception task.
}
  }
  \label{fig:re2st1}
\end{figure*}

\begin{figure*}[htb]
  \centering
\includegraphics[width=1.0\linewidth]{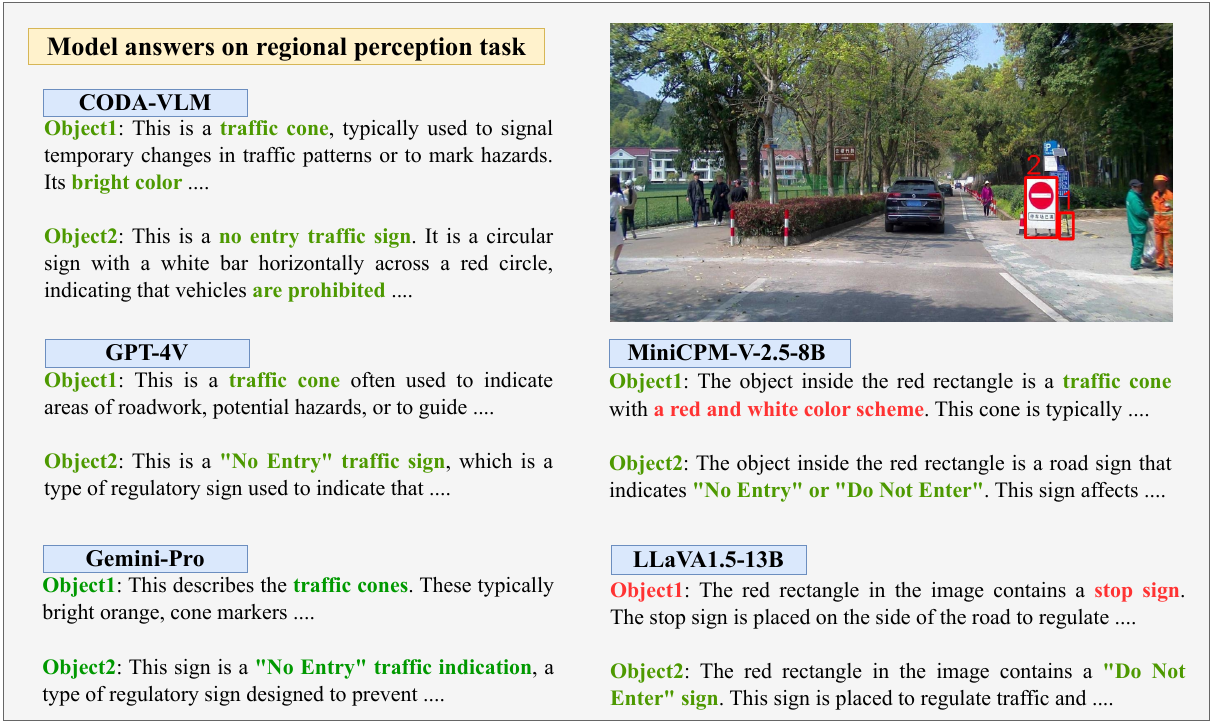}
  \caption{
  \textbf{
   More qualitative comparison for the regional perception task. }
  }
  \label{fig:re1st2}
\end{figure*}

\begin{figure*}[htb]
  \centering
\includegraphics[width=1.0\linewidth]{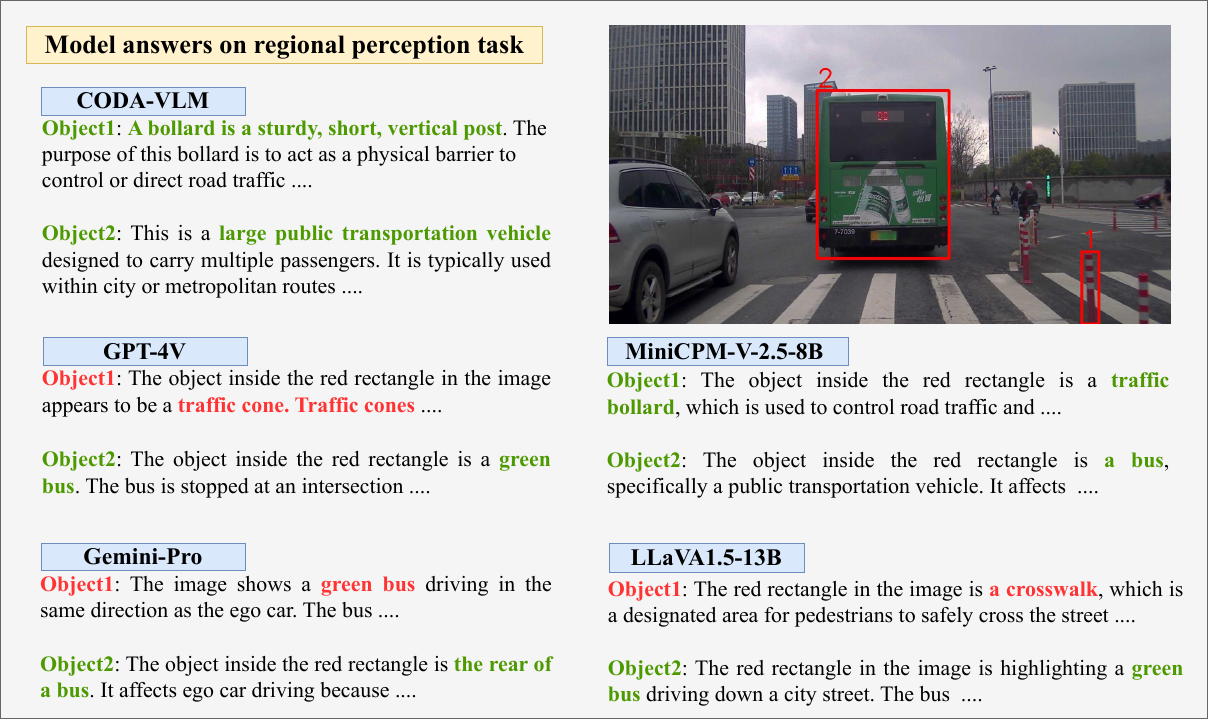}
  \caption{
  \textbf{
  More qualitative comparison for the regional perception task.  }
  }
  \label{fig:re2st2}
\end{figure*}

\begin{figure*}[htb]
  \centering
\includegraphics[width=1.0\linewidth]{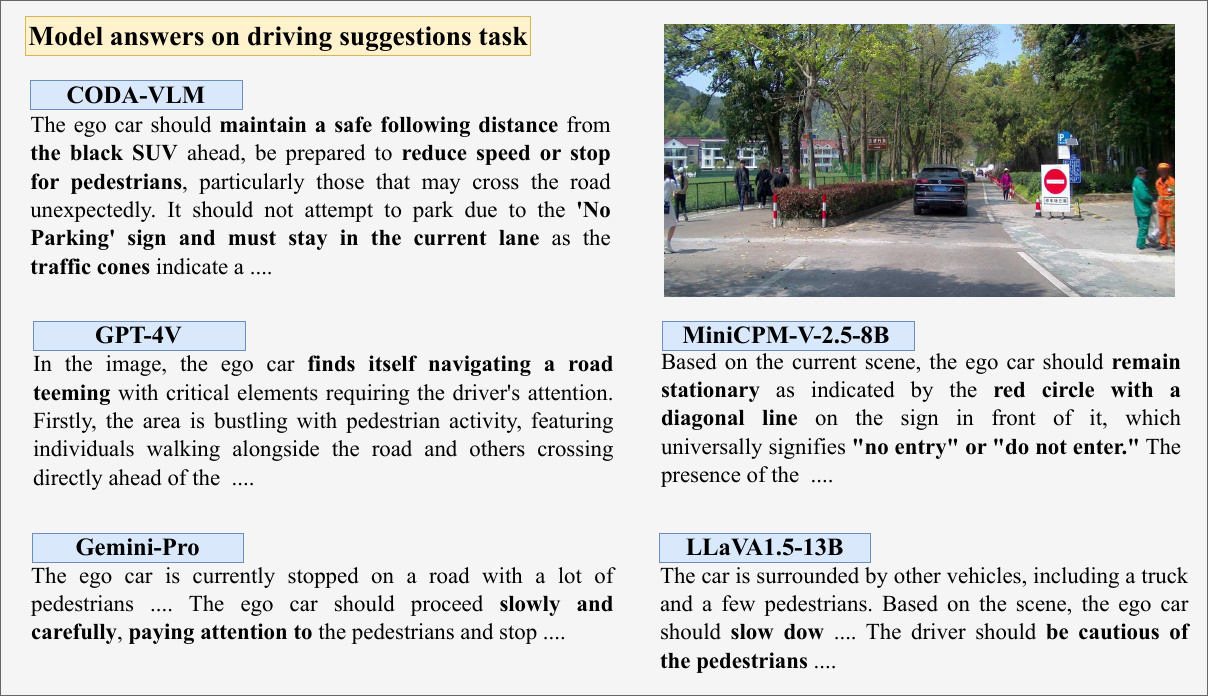}
  \caption{
  \textbf{
  More qualitative comparison for the driving suggestions task.}
  }
  \label{fig:re1st3}
\end{figure*}

\begin{figure*}[htb]
  \centering
\includegraphics[width=1.0\linewidth]{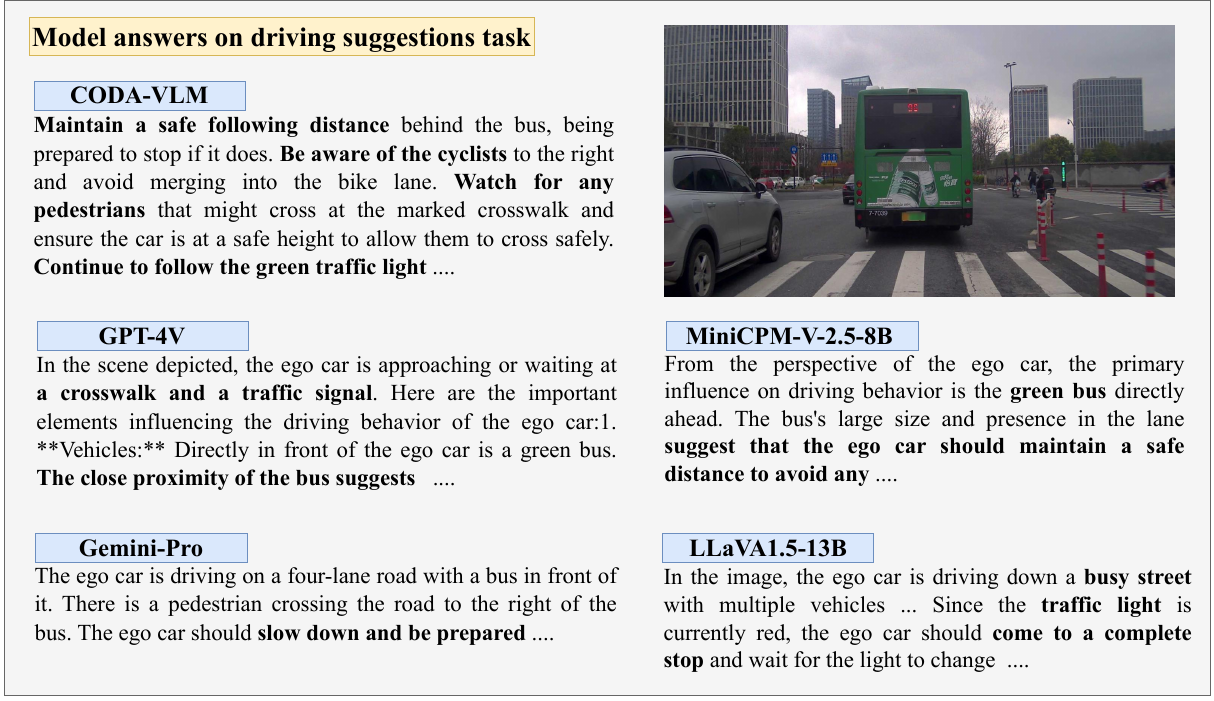}
  \caption{
  \textbf{
  More qualitative comparison for the driving suggestions task. }
  }
  \label{fig:re2st3}
\end{figure*}

In this section, we present three data examples from CODA-LM, as illustrated in Figures~\ref{fig:ov1} to~\ref{fig:ov3}. 
Building on CODA-LM, we subsequently analyze the responses from different LVLMs across three tasks, as shown in Figures~\ref{fig:re1st1} to~\ref{fig:re2st3}.

\Repeat{20}{\vphantom{0}\\}


\begin{table*}[t]
    \centering
    \scalebox{1.0}{
    \begin{tabular}{@{}ll| cccc@{}}
        \toprule
        \multirow{2}{*}{\textbf{Source}} & \multirow{2}{*}{\textbf{Model}} & \multicolumn{4}{c}{
        \textbf{Metrics $\uparrow$}} \\
        \cmidrule(lr){3-6}
        
        &  & \textbf{BLEU4} & \textbf{METEOR} & \textbf{CIDEr} & \textbf{SPICE} \\
        \midrule
        \multirow{4}{*}{Open} & MiniGPT-v2-7B~\cite{chen2023minigpt} & 0.6  & 5.3  & 0.6   & 4.4 \\
        & Shikra-7B~\cite{chen2023shikra}  & 1.5  & 8.7 &  0.0  & 5.2 \\
        & LLaVA1.5-7B~\cite{liu2023improved}  & 1.9  & 13.9 &  0.9  & 9.8  \\
        &LLaVA1.5-13B~\cite{liu2023improved}  & \textbf{2.7}  & 16.0 &  1.1  & 13.9  \\
        \midrule
        \multirow{2}{*}{Commercial} & Gemini Pro~\cite{gemini} & 1.9  & 12.9  & \textbf{4.8}   & 16.0 \\
        & GPT-4V~\cite{gpt4v} & 2.3  & \textbf{17.4}  & 0.0   & \textbf{19.2} \\
        \bottomrule
    \end{tabular}
    }
    \vspace{-3mm}
    \caption{\textbf{Comparison on regional perception using traditional evaluation metrics.}
    Although efficient, traditional metrics can hardly reflect the capabilities of LVLMs and differentiate models with different abilities, especially for complicated tasks like autonomous driving. By default, we adopt the \texttt{Text-Score} as the primary metric.
    }
    \vspace{-5mm}
    \label{tab:ablation_suggest}
\end{table*}

